%% file: conference_101719.tex
\documentclass[conference]{IEEEtran}
\IEEEoverridecommandlockouts
\usepackage{cite}
\usepackage{comment}
\usepackage{amsmath,amssymb,amsfonts}
\usepackage{algorithm}
\usepackage{algpseudocode}
\usepackage{graphicx}
\usepackage{textcomp}
\usepackage{xcolor,enumitem}

\def\BibTeX{{\rm B\kern-.05em{\sc i\kern-.025em b}\kern-.08em
    T\kern-.1667em\lower.7ex\hbox{E}\kern-.125emX}}

\input{mathcommand}

\usepackage{amsthm}
\newtheorem{theorem}{Theorem}
\newtheorem{proposition}{Proposition}
\newtheorem{lemma}{Lemma}

\newcommand{\ac}{\text{A}}

\newcommand{\regr}{\text{Reg}}

\begin{document}

\title{Bi-Level Online Provisioning and Scheduling with Switching Costs and Cross-Level Constraints}

\author{\IEEEauthorblockN{Jialei Liu}
\IEEEauthorblockA{\textit{Dept. of Electrical Engineering} \\
\textit{University at Buffalo, Buffalo, NY} \\
jialeili@buffalo.edu}
\and
\IEEEauthorblockN{C. Emre Koksal}
\IEEEauthorblockA{\textit{Dept. of Electrical and Computer Engineering} \\
\textit{The Ohio State University, Columbus, OH} \\
koksal.2@osu.edu}
\and
\IEEEauthorblockN{Ming Shi}
\IEEEauthorblockA{\textit{Dept. of Electrical Engineering} \\
\textit{University at Buffalo, Buffalo, NY} \\
mshi24@buffalo.edu}
}

\maketitle

\begin{abstract}
We study a bi-level online provisioning and scheduling problem motivated by network resource allocation, where provisioning decisions are made at a slow time scale while queue-/state-dependent scheduling is performed at a fast time scale. We model this two-time-scale interaction using an upper-level online convex optimization (OCO) problem and a lower-level constrained Markov decision process (CMDP). Existing OCO typically assumes stateless decisions and thus cannot capture MDP network dynamics such as queue evolution. Meanwhile, CMDP algorithms typically assume a fixed constraint threshold, whereas in provisioning-and-scheduling systems, the threshold varies with online budget decisions. To address these gaps, we study bi-level OCO-CMDP learning under switching costs (budget reprovisioning/system reconfiguration) and cross-level constraints that couple budgets to scheduling decisions. We build an algorithm to solve this bi-level problem by developing several new techniques, including a carefully designed dual feedback that returns the budget multiplier as sensitivity information for the upper-level update and a lower level that solves a budget-adaptive safe exploration problem via an extended occupancy-measure linear program. We establish near-optimal regret and high-probability satisfaction of the cross-level constraints.
\end{abstract}


\section{Introduction}\label{sec:introduction}

\begin{figure}[t]
  \centering
\includegraphics[width=0.49\textwidth]{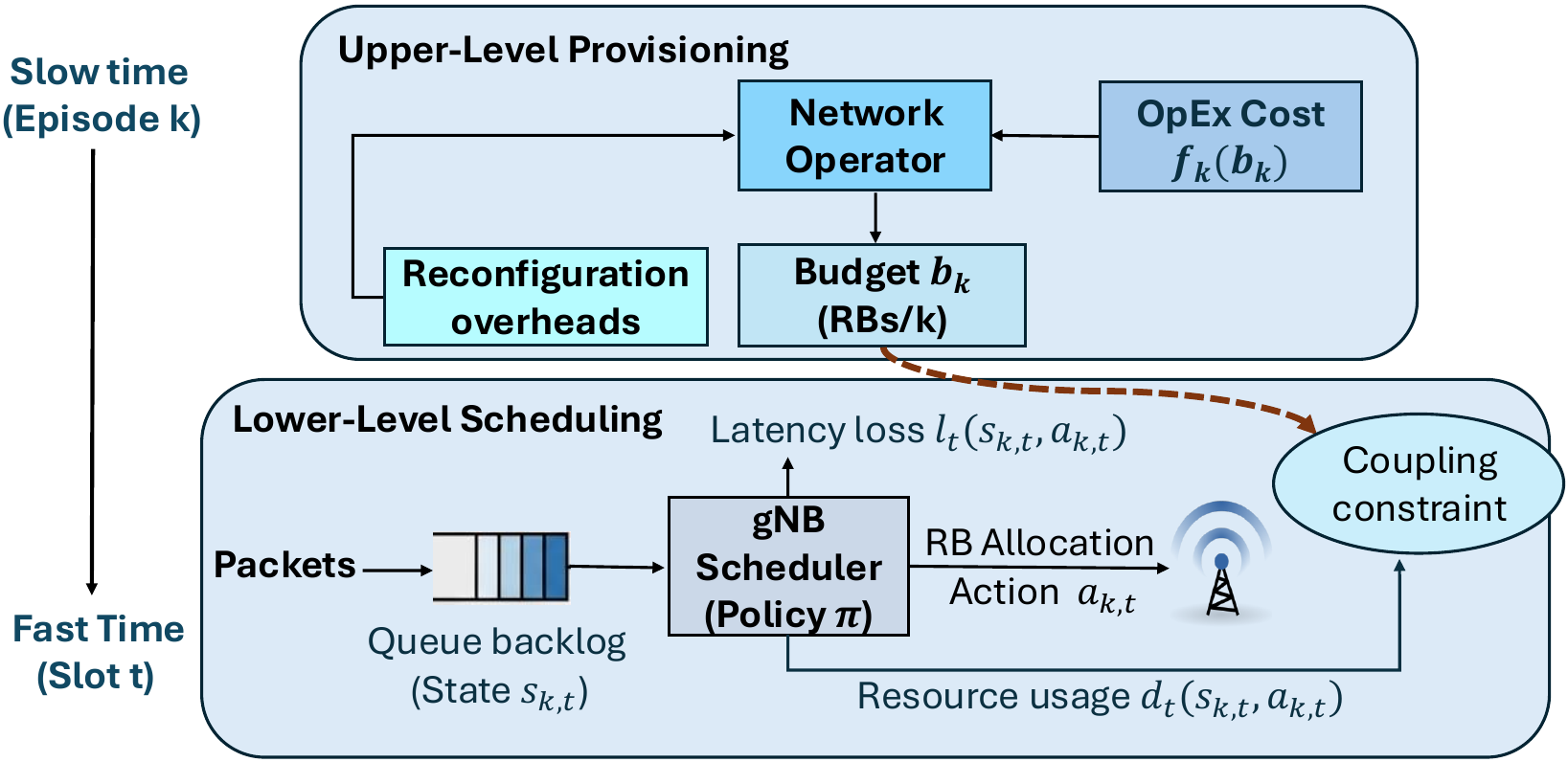}
  \caption{Online bandwidth provisioning with queue-aware scheduling.}
  \label{fig:motivating}
\end{figure}

Modern networking systems, such as 5G/6G network and edge computing, must support heterogeneous services (e.g., video streaming, augmented reality, virtual reality, and autonomous driving) over shared wireless infrastructure~\cite{lin2022leveraging}. A central operational challenge in these systems is online resource provisioning under uncertainty. In particular, a network operator must decide how much bandwidth/compute/energy to allocate to a slice over time, balancing quality of service (QoS) against operational expenditure (OpEx). Practical systems exhibit \emph{multiple coupled time scales}. As illustrated in our motivating example online bandwidth provisioning with queue-aware scheduling (see Fig.~\ref{fig:motivating}), the operator provisions an episode-level resource budget at a slow time scale (e.g., bandwidth reserved for a slice over seconds/minutes). This budget directly incurs OpEx and influences the QoS that can be achieved through scheduling. Within each episode, the scheduler performs state-dependent, packet-level actions at a fast time scale (e.g., allocating resource blocks every millisecond). These actions are \emph{constrained} by the provisioned resource budget and aim to reduce latency and congestion, improving QoS \cite{lin2006tutorial,le2009delay}. Moreover, changing the provisioning level across episodes is not free. It induces reconfiguration overheads (e.g., hardware retuning, virtualization/migration cost, and instability), which are naturally captured by \emph{switching costs} \cite{shi2021competitive,li2020online}. See Sec.~\ref{sec:motivating_example} for details.

Online convex optimization (OCO) is a classical abstraction for sequential resource allocation and has been extensively studied under adversarial and stochastic settings \cite{zinkevich2003online,shi2021competitive,li2020online}. Recent work incorporates switching costs and related operational constraints to penalize aggressive changes in decisions, which is attractive for network slicing. However, OCO with switching costs typically does not model endogenous Markovian state evolution nor provide safety under unknown state dynamics. For example, the per-round loss depends on the current decision and does not capture the evolution of system states, while in networked systems, state variables such as queue lengths evolve endogenously. Consequently, provisioning decisions that appear cost-effective in the short term may cause persistent congestion and latency violations.

Stateful scheduling problems are naturally modeled as Markov decision processes (MDPs), and a rich literature develops provably efficient reinforcement learning (RL) algorithms~\cite{sutton1998reinforcement,shi2023nearsafe,roknilamouki2025provably}. Yet, existing constrained MDP (CMDP) works typically assume that the budget, i.e., the constraint threshold, is \emph{fixed and exogenously specified}~\cite{shi2023nearsafe,roknilamouki2025provably}. However, in practice, the operator must learn what and how much budget to provision, while accounting for switching overhead. Thus, the constraint threshold itself is a primary decision variable.

Moreover, hierarchical RL (HRL) is a widely studied learning framework for multi-time-scale decision making \cite{barto2003recent}. However, its objective structure differs fundamentally from our resource provisioning problem. HRL typically optimizes a single, unified reward through task decomposition. In contrast, our problem involves a bi-level trade-off between provisioning cost  and service performance. Moreover, HRL generally allows frequent switching between sub-policies.

Motivated by the above gaps, we propose a bi-level online optimization and learning framework that integrates \emph{online budget provisioning} with \emph{state-aware safe scheduling} (without violating cross-level budget constraints). At the upper level, an OCO operator optimizes and provisions budgets $\{b_k\}_{k=1}^K$ across episodes, incurring a convex service cost and a quadratic switching cost. At the lower level, the system evolves as an episodic CMDP under \emph{varying} thresholds. Given the optimized provisioning budget $b_k$, the agent must learn a safe policy via RL under unknown dynamics. The two levels are coupled through an episode-wise budget constraint and an episode-wise objective that aggregates cross-level costs.

Designing an online algorithm for this coupled problem is technically challenging. \emph{(i) The upper-level effective optimization loss is implicit and non-stationary.} Beyond the unknown convex service cost and switching penalty, it depends on the unknown optimal performance of the lower-level controller. To address this, we leverage a carefully designed dual variable of the budget constraint counterpart to construct an approximated subgradient for evaluating the upper-level sensitivity. \emph{(ii) The lower-level CMDP faces a varying constraint threshold, since budget $b_k$ changes.} This renders standard safe/constrained RL methods with fixed constraints inapplicable. We address this by designing a budget-adaptive
exploration algorithm. In summary, our main contributions are as follows.
\begin{itemize}[leftmargin=*]
    \item We propose a new bi-level online optimization and learning algorithm for provisioning and scheduling with budget switching costs and cross-level budget constraints. At the upper level, we leverage dual multipliers returned by the lower-level (interpretable as the marginal value of extra budget), yielding a subgradient signal despite unknown stateful dynamics. At the lower level, we solve the episode-varying (decision-dependent) budget via budget-adaptive safe exploration and an extended occupancy-measure linear programming (LP), which both enforces feasibility and produces the dual feedback needed to couple the two levels.
    \item We establish near-optimal regret bounds with high-probability safety guarantees. Our analysis introduces a bi-level regret decomposition that explicitly quantifies how lower-level model-estimation and dual-signal errors propagate into the upper-level OCO updates, and it leverages the extended-LP strong duality to couple constraint satisfaction with sensitivity-based budget learning under switching costs.
    \item We instantiate the framework in a wireless network scenario, modeling the coupling between episode-level bandwidth provisioning and queue scheduling. Compared to decoupled baselines, our approach achieves a balance between cost and latency while maintaining strict physical feasibility.
\end{itemize}

\emph{Notations:} $\Delta(\mathcal{X})$ denotes the probability simplex over a given set $\mathcal{X}$. The operator $a \vee b$ denotes $\max\{a,b\}$. We use $[T]=\{1,\dots,T\}$ and $[x]^+ = \max\{0,x\}$.

\section{Problem Formulation}\label{sec:problemformulation}

\begin{figure}[t]
  \centering
  \includegraphics[width=0.50\textwidth]{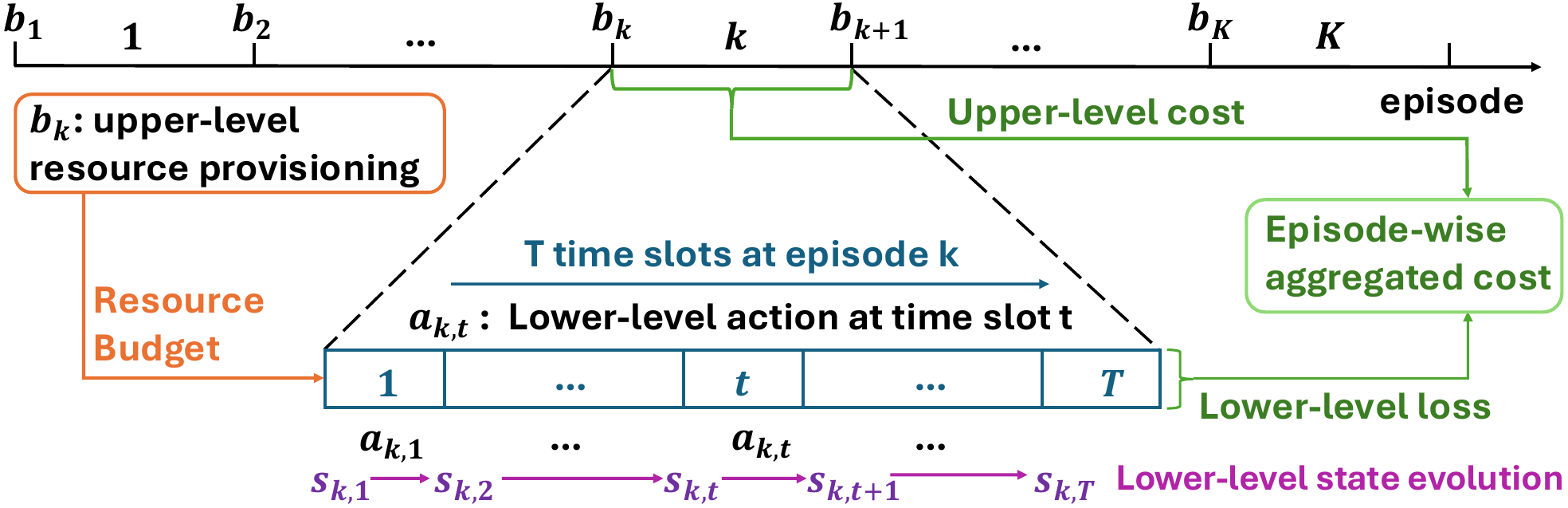}
  \caption{Coupled two-time-scale decision making: at the beginning of episode $k$, the agent provisions a resource budget $b_k$; within the episode, it follows a state-dependent policy selecting actions $a_{k,t}$ over $T$ time slots, incurring losses that aggregate to the episode cost and resource usage  that must satisfy the cross-level budget constraint.}
  \label{fig:timescale}
\end{figure}

We study a sequential decision-making problem with \emph{two coupled time scales}. An agent interacts with an environment over $K$ episodes (slow time scale). Within each episode, the environment evolves over $T$ time slots (fast time scale). In each episode $k\in[K]$, the agent chooses an \emph{upper-level} budget decision $b_k$ once per episode and a \emph{lower-level} sequence of state-dependent actions $\{a_{k,t}\}_{t=1}^T$ within the episode. The two levels are coupled through an episode-wise budget constraint (relating $b_k$ to $\{a_{k,t}\}_{t=1}^T$) and an episode-wise objective aggregating upper-level and lower-level costs, as
depicted in Fig.~\ref{fig:timescale}.

\subsection{Upper-Level Budget Provisioning}\label{subsec:modelupperlevel}

At the slow time scale, at the beginning of each episode $k\in[K]$, the agent selects an upper-level budget $b_k \in \mathcal{B}\triangleq [B_{0},T]$, which can represent the episode-level resource budget provisioned for a network slice (e.g., bandwidth) over seconds/minutes. Choosing $b_k$ incurs an episode-dependent convex service cost $f_k(b_k)$, which is \emph{not} revealed before the decision is made and is revealed at the end of episode $k$. As is typical in online convex optimization with switching costs~\cite{li2020online,hazan2016introduction,shi2021competitive}, the service cost $f_k(\cdot)$ could change arbitrarily, and it is $\theta_f$-strongly convex and differentiable on $\mathcal{B}$ with bounded gradients $\|\nabla f_k(\cdot)\|\le F$.

In addition, we consider the switching cost, i.e., changing the budget across episodes incurs an additional cost $\alpha \|b_k-b_{k-1}\|^2$, where $b_0=0$, and $\alpha>0$ quantifies reconfiguration overhead (e.g., signaling and control-plane costs, hardware retuning delays, and virtualization/migration overhead).

\subsection{Lower-Level Stateful Scheduling}\label{subsec:modellowerlevel}

At the fast time scale, within each episode, the environment evolves over $t=1,\ldots,T$ time slots according to a finite-horizon (possibly time-inhomogeneous) MDP $\mathcal{M}=(\mathcal{S},\mathcal{A},T,\{l_t\}_{t=1}^T,\{d_t\}_{t=1}^T,\{P_t\}_{t=1}^T)$. Here, $\mathcal{S}$ and $\mathcal{A}$ are the state and action spaces, $l_t:\mathcal{S}\times\mathcal{A}\to[0,1]$ is the stage loss, $d_t:\mathcal{S}\times\mathcal{A}\to[0,1]$ is the per-step resource consumption, and $P_t(\cdot\mid s,a)\in\Delta(\mathcal{S})$ is the \emph{unknown} transition kernel.

At each time $t$ of episode $k$, the agent observes the current state $s_{k,t}$ (e.g., queue backlogs), and then selects a (possibly randomized) scheduling/allocation action $a_{k,t}\sim \pi_{k,t}(\cdot\mid s_{k,t})$, where $\pi_k^{\ac}=(\pi_{k,t})_{t=1}^T$ is a policy with $\pi_{k,t}:\mathcal{S}\to\Delta(\mathcal{A})$. The agent then incurs a stage loss $l_t(s_{k,t},a_{k,t})$ (e.g., delay/backlog penalty), consumes resources $d_t(s_{k,t},a_{k,t})$ (e.g., used resource blocks), and the system evolves to next state $s_{k,t+1}\sim P_t(\cdot\mid s_{k,t},a_{k,t})$ due to unknown arrivals and service dynamics.

For compactness, we define the expected cumulative loss and consumption under policy $\pi$
and model $P$ as follows,
\begin{align}
V_{r}(\pi;P) & \triangleq \E_{\pi,P}\left[ \sum\nolimits_{t=1}^T r_t(s_t,a_t) \right], \label{eq:Vl_def}
\end{align}
where $r\in\{l,d\}$, and the expectation is taken with respect to the randomness of actions under $\pi$ and states under $P$.

\subsection{Cross-Level Coupling}\label{subsec:modelleveldependency}

The decisions in the two levels are coupled as follows.

\subsubsection{\textbf{Coupling 1 (Cross-Level Budget Constraint)}} Given upper-level budget $b_k$, the lower-level policy executed in episode $k$ must satisfy the episode-wise budget constraint, i.e.,
\begin{align}
V_d(\pi_k^{\ac};P) \le b_k. \label{eq:cross_level_constraint}
\end{align}
We define the feasible policy set under budget $b$ as
\begin{align}\label{safe_policy}
\Pi(b) \triangleq \{\pi^{\ac}: V_d(\pi^{\ac};P)\le b\}.
\end{align}
Then, (\ref{eq:cross_level_constraint}) is equivalently $\pi_k^{\ac} \in \Pi(b_k)$. This is a \emph{cross-level} constraint because its left-hand side is determined by lower-level actions while its right-hand side is the upper-level budget.

\subsubsection{\textbf{Coupling 2 (Cross-Level Joint Cost)}} Given $(b_k,\pi_k^{\ac})$, we define the episode cost as follows,
\begin{align}
C_k(b_k,\pi_k^{\ac}) \triangleq f_k(b_k) + \alpha\|b_k-b_{k-1}\|^2 + \beta V_l(\pi_k^{\ac};P), \label{eq:episode_total_cost}
\end{align}
where $\beta>0$ weighs within-episode performance relative to provisioning and reconfiguration costs. The first two terms capture upper-level operating and switching overhead, and the last term captures the cumulative loss induced by stateful allocation/control within the episode, subject to $\pi_k^{\ac}\in\Pi(b_k)$.

\subsection{Online Optimization Objective and Challenges}\label{subsec:goal_metric}

The agent aims to minimize the cumulative cross-level joint cost over $K$ episodes under the cross-level constraints, i.e.,
\begin{subequations}\label{eq:overallproblem}
\begin{align}
\min\nolimits_{\{b_k,\pi_k^{\ac}\}_{k=1}^K}\quad & \sum\nolimits_{k=1}^K C_k(b_k,\pi_k^{\ac}) \label{eq:overallproblemobjective} \\
\text{s.t.}\quad & b_k\in\mathcal{B}, \pi_k^{\ac}\in\Pi(b_k), \text{ for all } k\in[K].
\label{eq:overallproblemconstraint}
\end{align}
\end{subequations}
Solving (\ref{eq:overallproblem}) online is non-trivial due to the two couplings. First, the upper-level decision must be made without knowing $f_k$ in advance and its effective loss depends on the (learned) lower-level performance, since the value induced by $\pi_k^{\ac}$ is only realized through stateful dynamics with unknown transitions. Second, the lower level faces a constraint whose threshold is episode-dependent (through $b_k$), and thus requires safety guarantees that remain valid uniformly over the varying budget. Hence, traditional CMDP cannot be applied anymore.

\subsection{Performance Metric}\label{subsec:perf_metric}

We evaluate an online algorithm using the static regret, which compares against the best fixed budget and fixed lower-level policy in hindsight. Define the best static optimal solution
\begin{align}
(b^*,\pi^{\ac,*}) \in \argmin\nolimits_{b\in\mathcal{B},\pi^{\ac}\in\Pi(b)} \sum\nolimits_{k=1}^K C_k(b,\pi^{\ac}), \label{eq:static_benchmark_def}
\end{align}
where $\pi^{\ac,*}$ could be non-stationary in an episode, i.e., $\pi^{\ac,*}=(\pi_t^*)_{t=1}^T$. The static regret after $K$ episodes is then defined as
\begin{align}
\regr(K,b,\pi) \triangleq \sum\nolimits_{k=1}^K [C_k(b_k,\pi_k^{\ac}) - C_k(b^*,\pi^{\ac,*})]. \label{eq:regret-def}
\end{align}
A sublinear bound $\regr(K, b,\pi)=o(K)$ implies that the average performance of the online algorithm converges to that of the best fixed budget-policy pair in hindsight.

\section{Motivating Example}\label{sec:motivating_example}

The formulation in Sec.~\ref{sec:problemformulation} captures network settings where a \emph{slow} provisioning decision must be coordinated with \emph{fast}, MDP-dependent scheduling, e.g., queue-aware stochastic network control and scheduling~\cite{lin2006tutorial,le2009delay}, distributed low-complexity scheduling under interference constraints~\cite{gupta2009low}, and delay-sensitive wireless scheduling with queue information~\cite{ji2012delay}. We instantiate the framework in \textbf{online bandwidth provisioning with queue-aware scheduling}, matching the abstraction in Fig.~\ref{fig:motivating} and the two-time-scale timeline in Fig.~\ref{fig:timescale}.

\paragraph{Scenario (two time scales)}
Consider a next-generation NodeB (gNB) serving an ultra-reliable low-latency communications (URLLC) slice with stochastic packet arrivals in a radio access network (RAN). Slice provisioning/configuration is updated over coarse control intervals (seconds/minutes), while the medium access control (MAC) scheduler allocates resource blocks (RBs) at fine granularity (milliseconds) in response to queue and channel fluctuations.

\paragraph{Upper level (provisioning with switching overhead)} At the beginning of episode $k$, the operator provisions a bandwidth budget $b_k$, e.g., total RB$\cdot$slot units over $T$ slots. Provisioning incurs an operating cost $f_k(b_k)$ (spectrum leasing, activation energy, or opportunity cost relative to other slices), which may vary with network conditions and is revealed only after the episode. Reconfiguring the provisioned bandwidth across episodes induces overhead, modeled by $\alpha\|b_k-b_{k-1}\|^2$.

\paragraph{Lower level (queue-aware scheduling under a budget)} Within episode $k$, over slots $t=1,\ldots,T$, the scheduler observes the state $s_{k,t}$ (e.g., slice queue backlog) and selects an action $a_{k,t}$ (RB allocation/transmission decision). The action incurs a stage loss $l_t(s_{k,t},a_{k,t})$ (delay/backlog penalty reflecting QoS) and consumes resources $d_t(s_{k,t},a_{k,t})$ (RB usage), while the queue evolves via arrivals and service dynamics. This statefulness is essential, since insufficient service can accumulate backlog and trigger persistent latency violations.

\paragraph{Online optimization and learning with cross-level coupling} The provisioned budget constrains scheduling through the episode-wise budget constraint~(\ref{eq:cross_level_constraint}), requiring the expected cumulative RB usage within episode to not exceed budget. Hence, the upper level must adapt $b_k$ online under switching costs, while the performance of a given $b_k$ is realized through a lower-level CMDP scheduler under unknown dynamics.

\section{Bi-Level Algorithm Design}\label{sec:algorithmdesign}

We design an online algorithm for the problem in Sec.~\ref{sec:problemformulation}.

\subsection{Upper Level OCO via Lower Level Dual Feedback}\label{subsec:upper_oco}

\begin{algorithm}[t]
\caption{Bi-Level Optimization and Learning (BLOL)}
\label{alg:upper_level_framework}
\begin{algorithmic}[1]
\State \textbf{Initialization:} $b_1\gets B_{0}$; history $\mathcal{H}\gets\emptyset$.
\For{$k=1$ to $K$}
    \State Execute BALDE (Algorithm~\ref{alg:bdope_subroutine}) in episode $k$ under budget $b_k$ and obtain $(\pi_k,\lambda_k,\mathcal{H})$.
    \If{$k\le K_0$}
        \State $b_{k+1}\gets B_{0}$
    \Else
        \State Compute approximate subgradient according to (\ref{eq:approx_subgrad}) and update resource budget according to (\ref{eq:upper_update}).
    \EndIf
\EndFor
\end{algorithmic}
\end{algorithm}

The upper-level selects a budget sequence $\{b_k\}_{k=1}^K$ to provision episode-wise resources while accounting for the unknown $f_k$ and switching overhead. Moreover, the effective loss of $b_k$ depends on the \emph{best achievable lower-level performance} under the provisioned budget. This relationship is implicit because the operator lacks an analytical model of the packet arrival and scheduling dynamics, i.e., unknown MDP dynamics. We address these challenges as follows and see Algorithm~\ref{alg:upper_level_framework}.

\subsubsection{Reduce bi-level coupling to an ``oracle'' effective loss} To isolate the upper-level design, we first analyze an \emph{idealized} setting in which the lower-level scheduling policy can be optimized for any budget $b_k$. Define the corresponding cost
\begin{align}
g_k(b_k) \triangleq f_k(b_k) + \beta L_k^{*}(b_k),
\label{eq:gk-def-rev}
\end{align}
where $f_k(b_k)$ is the upper-level service cost, and $L_k^{*}(b_k)$ is the optimal lower-level scheduling loss under budget $b_k$, i.e.,
\begin{align}
L_k^{*}(b_k) \triangleq \min\nolimits_{\pi^{\ac}\in \Pi(b_k)} \E_{\pi^{\ac},P} \left[ \sum\nolimits_{t=1}^T l_t(s_t,a_t) \right].
\label{eq:Lstar-def}
\end{align}
Under this \emph{fictitious} oracle
view, the upper level would be reduced to an ideal OCO problem over $\gB$, i.e.,
\begin{align}
\min\nolimits_{{b_{k}}\in\gB} \Big\{\sum\nolimits_{k=1}^K g_k(b_k) + \sum\nolimits_{k=1}^{K}\alpha \|b_k-b_{k-1}\|^2\Big\}.
\label{eq:upper_oracle_oco}
\end{align}

\subsubsection{Dual-based subgradient approximation} Directly optimizing (\ref{eq:upper_oracle_oco}) is still challenging because $L_k^{*}(b_k)$ depends on the unknown lower-level MDP dynamics. Our key observation is that, by strong duality of the linear programing (LP) formulation of finite-horizon CMDPs, the optimal dual multiplier $\lambda_k^*(b_k)$ of the budget constraint $b_k$ provides the sensitivity of the optimal value with respect to the budget. Physically, $\lambda_k^*(b_k)$ signal the upper level how much QoS could improve if more RBs were allocated. Thus, $-\lambda_k^{*}(b_k)$ constitutes a valid subgradient of loss $L_k^{*}(b_k)$ (Please see Appendix \ref{F} for the complete proof of Lemma~\ref{lem:Lk-subgradient}). Specifically, for any $b_k\in\gB$, the subgradient of the true loss is given by
\begin{align}
\partial g_k(b_k) = \nabla f_k(b_k) + \beta \partial L_k^{*}(b_k) = \nabla f_k(b_k) - \beta \lambda_k^{*}(b_k), \nonumber
\end{align}
However, since the true MDP dynamics are unknown, the true multiplier $\lambda_k^*(b_k)$ is inaccessible. In our algorithm, we approximates it with the empirical dual estimate $\lambda_k$ returned by the lower-level scheduler. Then, the update direction is
\begin{align}
\hat h_k \triangleq \nabla f_k(b_k) - \beta \lambda_k.
\label{eq:approx_subgrad}
\end{align}
While $\hat h_k$ is an approximation of the  $\partial L_k^*(b_k)$, it acts as an exact subgradient for the \emph{surrogate objective} $\hat g_k(b_k) = f_k(b_k) + \beta \hat L_k(b_k)$, where $\hat L_k(b_k)$ is the surrogate lower-level optimal loss. This connection allows us to analyze the algorithm as performing projected subgradient descent on the surrogate functions $\{\hat g_k\}$, with the gap between true and surrogate objectives controlled by the lower-level learning guarantees, and it will appear in the upper-level regret bound.

\subsubsection{Switching cost control via stable OCO updates} While the total episode cost includes the switching term $\alpha\|b_k-b_{k-1}\|^2$, our update rule operates on (approximate) subgradients of $g_k$. As we show in Lemma~\ref{lem:oco_regret}, a decaying step size $\eta_k\propto 1/k$ yields stable budget trajectories with controlled path length, which in turn controls the cumulative switching costs without requiring a switching regularizer in the update.

\subsubsection{Projected subgradient descent with a warm-up phase} We adopt a warm-up period of $K_0=\tilde{\mathcal O}(S^2AT^4/\gamma^2)$ episodes to allow the lower-level learner to collect sufficient samples for dual estimates, and see the proof sketch of Lemma~\ref{thm:bdope_main}.
\begin{itemize}
    \item \emph{Warm-up ($k\le K_0$):} set $b_k=B_{0}$ and run the lower-level learner to explore the environment.
    \item \emph{Online update ($k>K_0$):} after observing $f_k(\cdot)$ and receiving $\lambda_k$ from the lower level, update the budget
    \begin{align}
    b_{k+1} = \Pi_{\gB} \left( b_k - \eta_k \hat h_k \right),
    \label{eq:upper_update}
    \end{align}
    where the step-size is $\eta_k = 1 / [\theta_g (k-K_0)]$.
\end{itemize}

\subsection{Lower Level RL Under Upper Level Dynamic Budget}\label{subsec:lower_cmdp}

\begin{algorithm}[t]
\caption{Budget-Adaptive Learning with Dual Estimation}
\label{alg:bdope_subroutine}
\begin{algorithmic}[1]
\State \textbf{Input:} episode $k$, budget $b_k$, history $\mathcal{H}$, confidence level $\delta$, baseline policy $\pi_{\mathrm{base}}$.
\State Compute $\hat P_{k,t}$ and $\beta^p$ using $\mathcal{H}$ according to (\ref{model}-\ref{eq:beta_p}).
\State Calculate estimates $\bar l_{k,t}$ and $\bar d_{k,t}$ according to (\ref{eq:dbar}-\ref{eq:lbar}).

\If{$k \le K_0$}
    \State $\pi_k^{\ac} \gets \pi_{\mathrm{base}}$ \Comment{Use safe baseline during warm-up}
    \State $\lambda_k \gets 0$ \Comment{Dual variable is not used in warm-up}
\Else
    \State Solve the extended LP (\ref{eq:ext-LP}) with budget $b_k$, and obtain optimal policy $\pi_k^{\ac}$ and dual multiplier $\lambda_k$ of the constraint.
\EndIf

\For{$t = 1$ to $T$}
    \State Observe state $s_{k,t}$, next execute action $a_{k,t} \sim \pi_{k,t}(\cdot|s_{k,t})$, and then observe loss $l_t(s_{k,t},a_{k,t})$, consumption $d_t(s_{k,t},a_{k,t})$, and next state $s_{k,t+1}$.
    \State Update history: $\mathcal{H} \gets \mathcal{H} \cup \{(s_{k,t}, a_{k,t}, s_{k,t+1})\}$.
\EndFor
\State \textbf{Output:} $(\pi_k,\lambda_k,\mathcal{H})$.
\end{algorithmic}
\end{algorithm}

In episode $k$, given the budget $b_k$, the lower-level scheduler seeks a policy $\pi_k$ (for brevity, we write $\pi_k$ for $\pi_k^{\ac}$ in this subsection) that minimizes the expected cumulative loss $V_l(\pi_k;P)$ while satisfying the episode budget constraint $V_d(\pi_k;P) \le b_k$.

To guarantee safety \emph{uniformly} for dynamic constraint threshold and return a dual variable $\lambda_k$, we propose \emph{Budget-Adaptive Learning with Dual Estimation} (BALDE), see Algorithm~\ref{alg:bdope_subroutine}.

\subsubsection{Empirical estimates and confidence sets} We focus on uncertainty in the packet arrival and channel dynamics (transition probabilities), and the results can be extended to unknown $l_t$ and $d_t$ via standard concentration arguments. For each state-action pair $(s,a)$ at time sot $t$, define the visitation counts prior to episode $k$ as $n_{k,t}(s,a) = \sum\nolimits_{k'=1}^{k-1} \mathbf{1}\{s_{k',t}=s,~a_{k',t}=a\}$, $n_{k,t}(s,a,s') = \sum\nolimits_{k'=1}^{k-1} \mathbf{1}\{s_{k',t}=s,~a_{k',t}=a,~s_{k',t+1}=s'\}$. At the start of episode $k$, the empirical transition estimate is
\begin{align}\label{model}
\hat P_{k,t}(s'|s,a) = n_{k,t}(s,a,s') / [n_{k,t}(s,a)\vee 1],
\end{align}
Then, we construct a confidence set $\mathcal{P}_k$ centered at $\hat P_{k,t}$:
\begin{align}
\gP_k \triangleq \bigcap\nolimits_{t\in[T],\,s\in\sS,\,a\in\sA} \gP_{k,t}(s,a),
\end{align}
where, based on the empirical Bernstein inequality~\cite{maurer2009empirical}, $\gP_{k,t}(s,a) = \{P' : \big|P'_{t}(s'|s,a) - \hat P_{k,t}(s'|s,a)\big| \le \beta^{p}_{k,t}(s,a,s') \text{ for all } s' \}$, with radius
\begin{align}\label{eq:beta_p}
\textstyle \beta^{p}_{k,t}(s,a,s') = \sqrt{\frac{4\,\mathrm{Var}(\hat P_{k,t}(s'|s,a))\,L'}{n_{k,t}(s,a)\vee 1}} + \frac{14L'/3}{n_{k,t}(s,a)\vee 1},
\end{align}
and $L'=\log \big(\frac{2SATK}{\delta}\big)$. With probability at least $1-2\delta$, the true model $P$ lies in $\mathcal{P}_k$ for all $k$ (Please see Appendix \ref{thm:bdope_main_1} for the complete proof of Lemma~\ref{thm:bdope_main}).

\subsubsection{Pessimism for safety} To ensure the scheduled resources do not exceed the budget $b_k$ under model uncertainty, we inflate the constraint cost using a pessimistic penalty, i.e.,
\begin{align}
& \bar d_{k,t}(s,a) = d_{t}(s,a) + T \beta^{p}_{k,t}(s,a),
\label{eq:dbar}
\end{align}
where $\beta^{p}_{k,t}(s,a)\triangleq \sum\nolimits_{s'} \beta^{p}_{k,t}(s,a,s')$. $\bar d_{k,t}$ enforces a conservative constraint that
holds uniformly for all $P'\in\mathcal{P}_k$.

\subsubsection{Budget-aware optimism for efficient exploration} Pure pessimism can severely restrict exploration, especially for rarely visited state-action pairs. We therefore introduce an optimistic adjustment to the scheduling loss, i.e.,
\begin{align}
\bar l_{k,t}(s,a) = l_{t}(s,a) - [T^2\beta^{p}_{k,t}(s,a)] / (b_k-b_{\mathrm{base}}).
\label{eq:lbar}
\end{align}
The scaling by $(b_k-b_{\mathrm{base}})^{-1}$ is crucial, since when the current budget $b_k$ is tight (close to $b_{\mathrm{base}}$), the algorithm reduces aggressive exploration, and when $b_k$ is larger, the learner can explore more aggressively while still maintaining feasibility.

\subsubsection{Optimization under extended LP} Based on the construction above, we select the scheduling policy as follows,
\begin{align}\label{eq:dop_lower}
\pi_k = \argmin\nolimits_{\{\pi':V_{\bar d_k}(\pi';P') \le b_k;P'\in\gP_k\}} V_{\bar l_{k,t}}(\pi';P').
\end{align}
This formulation is optimistic in the objective (to learn low-latency scheduling) and pessimistic in the constraint (to ensure RB limits are respected). However, solving (\ref{eq:dop_lower}) efficiently and \emph{extracting the dual multiplier} associated with the budget constraint are essential and non-trivial in our bi-level design.

To this end, we cast (\ref{eq:dop_lower}) as an \emph{extended LP} in occupancy measures, which serves two roles: (i) it provides a tractable solver for the lower-level under time-inhomogeneous dynamics, and (ii) it yields the \emph{dual multiplier} estimate $\lambda_k$ for the budget constraint, which acts as a sensitivity signal (approximately $\partial_{b} L_k^{*}(b_k)$) required by the upper-level update. This primal-dual interface is critical not only for algorithm implementation but also for regret analysis, where $\lambda_k$ connects lower-level learning error to upper-level optimization error. Specifically, for a policy $\pi$ and transition kernels $P=\{P_t\}_{t=1}^T$, define the
state-action occupancy measure $w_t^{\pi}(s,a;P) \triangleq \Pr(s_t=s,a_t=a \mid \pi,P)$~\cite{shi2023near}. The expected cumulative loss and resource consumption are linear, i.e.,
\begin{align}
V_r^{\pi}(P) & = \sum\nolimits_{t=1}^T \sum\nolimits_{s,a} w_t^\pi(s,a;P) r_t(s,a) = \langle r, w^\pi \rangle, \label{eq:value_q}
\end{align}
where $r \in \{l,d\}$ and $w^\pi=\{w_t^\pi\}_{t=1}^T$. Any feasible occupancy measure must satisfy standard flow constraints. Let $\gW(P)$ denote the set of nonnegative $w^{\pi}$ satisfying, for all $s\in\mathcal{S}$,
\begin{align}
\sum\nolimits_{a} w_t^{\pi}(s,a) &= \sum\nolimits_{s',a'} P_{t-1}(s\mid s',a') w_{t-1}^{\pi}(s',a').
\label{eq:lp_flow}
\end{align}
Then, the CMDP under dynamic budget $b_k$ is equivalent to
\begin{align}
\min\nolimits_{w^{\pi}\in\gW(P)} \langle l, w^{\pi}\rangle, \text{ s.t. } \langle d, w^{\pi}\rangle \le b_k. \label{eq:lp_constraint}
\end{align}
An optimal policy can be obtained by normalization $\pi_t^*(a\mid s)=w_t^*(s,a) / \sum_{a'} w_t^*(s,a')$ for nonzero denominators. In BALDE, the transition model is unknown and is optimized within a confidence set $\gP_k$. To avoid bilinearity between $P'\in\gP_k$ and $w^{\pi}$, we introduce an augmented occupancy measure
\begin{align}
q_t^{\pi}(s,a,s') \triangleq w_t^{\pi}(s,a)\,P'_t(s'\mid s,a),
\end{align}
which represents the joint occupancy of $(s_t,a_t,s_{t+1})$. This yields a linear program in $q^{\pi}$ to represent the problem (\ref{eq:dop_lower}) :
\begin{subequations}\label{eq:ext-LP}
\begin{align}
\min\nolimits_{q^{\pi}} & \sum\nolimits_{t,s,a,s'} q_t^{\pi}(s,a,s') \bar l_t(s,a)
\label{eq:ext-LP-obj} \\
\text{s.t.} & \sum\nolimits_{t,s,a,s'} q_t^{\pi}(s,a,s') \bar d_t(s,a) \le b_k, \label{eq:ext-LP-constraint} \\
& \sum\nolimits_{a,s'} q_t^{\pi}(s,a,s') = \sum\nolimits_{s'',a''} q_{t-1}^{\pi}(s'',a'',s), \label{eq:ext-LP-flow} \\
& \hat P_{k,t}(s' \mid s,a) - \beta^p_{k,t}(s,a,s') \le \frac{q_t(s,a,s')}{\sum_{y} q_t(s,a,y)} \nonumber \\
& \le \hat P_{k,t}(s' \mid s,a)+\beta^p_{k,t}(s,a,s'), \forall s,a,s',t, \label{eq:ext-LP-conf} \\
& q_t(s,a,s')\ge 0, \forall s,a,s',t, \label{eq:ext-LP-nonneg}
\end{align}
\end{subequations}
where (\ref{eq:ext-LP-flow}) holds for all state $s$ and (\ref{eq:ext-LP-conf}) encodes the transition-dynamics confidence set constraint $P'_t(\cdot\mid s,a)\in\mathcal{P}_k$ in a linear form. Crucially, the dual variable associated with the budget constraint (\ref{eq:ext-LP-constraint}) is exactly the multiplier $\lambda_k$, which the lower level returns to the upper level as a subgradient and sensitivity feedback for updating the budget $b_k$.

\subsubsection{Initial safety under dynamic budgets} As is common in safe RL~\cite{shi2023nearsafe,roknilamouki2025provably}, we assume access to a baseline safe scheduler $\pi_{\mathrm{base}}$ (e.g., a conservative rate-limiting policy) satisfying $V_d(\pi_{\mathrm{base}};P)=b_{\mathrm{base}}$ with $b_{\mathrm{base}} < B_{0}$, which is used to guarantee feasibility in early learning episodes without any pre-samples. We only execute $\pi_{\mathrm{base}}$ during the warm-up phase.

\section{Regret Analysis and Budget Guarantee}\label{sec:regretanalysis}

In this section, we establish the regret upper bound and cross-level budget constraint satisfaction guarantee for our algorithm. The main technical challenges lie in quantifying the error quantification from the lower-level reinforcement learning under varying threshold to the upper-level optimization updates under coupling and estimated sensitivity feedback. We provide proof sketches for the lemmas and theorems (Please see complete proofs in our Appendix).

\subsection{Regret Decomposition}\label{regret_decomposition}

Recall the total regret after $K$ episodes $\regr(K, b,\pi) \triangleq \sum\nolimits_{k=1}^{K} C_k(b_k,\pi_k^{\ac}) - \sum\nolimits_{k=1}^{K} C_k(b^{*},\pi^{{\ac},*})$. For each episode $k$ and a fixed upper-level decision $b_k$, define the \emph{episode-wise optimal safe lower-level policy} under the true model $P$:
\begin{align}
\pi_k^{*}(b_k) \in \argmin\nolimits_{\pi\in\Pi(b_k)} V_l(\pi;P),
\label{eq:lower_oracle_def}
\end{align}
where $\Pi(b_k)$ is the feasible policy set induced by budget $b_k$. Using $\pi_k^{*}(b_k)$ as an intermediate comparator, we can decompose the final regret as follows,
\begin{align}
&\regr(K, b,\pi) = \underbrace{\sum\nolimits_{k=1}^{K} \Big( C_k(b_k,\pi_k^{\ac}) - C_k(b_k,\pi_k^{*}(b_k)) \Big)}_{\text{(I) Lower-level BALDE regret under budget $b_k$: } \regr_{\mathrm{LL}}(K, \pi)} \nonumber \\
&\quad + \underbrace{\sum\nolimits_{k=1}^{K} \Big( C_k(b_k,\pi_k^{*}(b_k)) - C_k(b^{*},\pi^{{\ac},*}) \Big)}_{\text{(II) Upper-level BLOL regret with switching costs: } \regr_{\mathrm{UL}}(K, b)}. \label{eq:regret_decomp}
\end{align}
Term (I) in (\ref{eq:regret_decomp}) represents the lower-level BALDE regret under varying budget $b_k$, i.e., the scheduling regret. By the definition of $C_k$ in (\ref{eq:episode_total_cost}), for a fixed budget $b_k$, the upper-level service cost and switching cost diminishes to zero and we have
\begin{align}
\regr_{\mathrm{LL}}(K,\pi) = \beta \sum\nolimits_{k=1}^{K}( V_l(\pi_k;P) - V_l(\pi_k^{*}(b_k);P) ).
\label{eq:LL_regret_simplify}
\end{align}
Thus, $\regr_{\mathrm{LL}}(K,\pi)$ is exactly the scaled learning regret of the lower-level algorithm BALDE under the budget sequence generated by the upper level. Term (II) in (\ref{eq:regret_decomp}) represents the upper-level BLOL regret with switching costs, i.e., the provisioning regret. Using the oracle effective loss $g_k(b)\triangleq f_k(b)+\beta L_k^{*}(b)$, where $L_k^{*}(b) = \min\nolimits_{\pi\in\Pi(b)} V_l(\pi;P)$ (see (\ref{eq:Lstar-def})), the upper-level regret term $\regr_{\mathrm{UL}}(K,b)$ becomes
\begin{align}
\sum\nolimits_{k=1}^K \left( g_k(b_k) - g_k(b^*) \right) + \alpha \sum\nolimits_{k=1}^K \|b_k - b_{k-1}\|^2.
\label{eq:oracle_cost_identity}
\end{align}

\subsection{Lower-Level Regret Bound}\label{subsec:LL_bound}

We first bound the lower-level CMDP scheduling regret.

\begin{lemma}[Lower-level BALDE regret]\label{thm:bdope_main}
Fix $\delta\in(0,1)$. With probability at least $1-3\delta$, executing BALDE (Algorithm~\ref{alg:bdope_subroutine}) under budget sequence $\{b_k\}_{k=1}^K$, we have ($\gamma = B_{0}-b_{\mathrm{base}}$)
\begin{align}
\sum\nolimits_{k=1}^{K} \Big( V_l(\pi_k;P) - V_l(\pi_k^{*}(b_k);P) \Big) \le \tilde{\mathcal O} \left(ST^{3}\sqrt{A K} / \gamma\right). \nonumber
\end{align}
\end{lemma}

Lemma~\ref{thm:bdope_main} provides a \emph{high-probability} regret guarantee for the lower-level scheduler under an \emph{episode-varying} budget sequence $\{b_k\}_{k=1}^K$, which is crucial for coupling with the upper-level provisioning decisions. The bound scales as $\tilde{\mathcal O}(\sqrt{K})$, i.e., the average per-episode suboptimality decays as $\tilde{\mathcal O}(1/\sqrt{K})$. This $\sqrt{K}$ dependence is optimal and matches the canonical statistical rate for learning under unknown Markov dynamics. The dependence on $\gamma$ makes explicit a fundamental \emph{safety-learning tradeoff}. $\gamma$ is the minimum safety slack that allows the algorithm to enforce a pessimistically inflated constraint (to guarantee feasibility under uncertainty) while still retaining room for exploration. When $\gamma$ is small, safe learners must behave conservatively, so a blow-up with $1/\gamma$ is unavoidable~\cite{shi2023nearsafe,roknilamouki2025provably}. Finally, the $T^3$ factor is characteristic of finite-horizon optimistic exploration with extended-LP-based planning. It reflects that estimation errors compound along length-$T$ trajectories and that additional factors appear when enforcing constraints pessimistically while maintaining optimism. Please see Appendix~\ref{thm:bdope_main_1} for the complete proof of Lemma~\ref{thm:bdope_main}.

\begin{proof}[Proof sketch of Lemma~\ref{thm:bdope_main}]
The proof follows an optimism-in-the-face-of-uncertainty argument together with pessimism for the constraint, but adapted to a time-varying budget $b_k$.

\subsubsection{Step 1 (High-probability event)} We construct confidence sets $\mathcal{P}_k$ around empirical transitions and define a good event $\mathcal{G}$ under which $P\in\mathcal{P}_k$ for all $k$. Empirical Bernstein bounds then imply $\Pr(\mathcal{G})\ge 1-3\delta$.

\subsubsection{Step 2 (Safety under a dynamic threshold)} We use a safe baseline policy $\pi_{\mathrm{base}}$ during warm-up phase $K_0$ to ensure feasibility when uncertainty is high in early exploration. For $k>K_0$, the solver ensures safety by incorporating a cumulative transition bonus $\epsilon_k^{\pi_k}(P_k) \triangleq T \mathbb{E}_{\pi_k, P_k} [\sum_{t=1}^T {\beta}^p_{k,t}]$, we have $V_d(\pi_k;P) \le V_d(\pi_k;P_k) + \epsilon_k^{\pi_k}(P_k) = V_{\bar{d}}(\pi_k;P_k)$. Since the solver ensures $V_{\bar{d}}(\pi_k;P_k) \le b_k$, it follows that $V_d(\pi_k;P) \le b_k$ under event $\mathcal{G}$ with probability at least $1-3\delta$.

\subsubsection{Step 3 (Regret bound)} We decompose the per-episode regret into two terms, i.e., $V_l({\pi_k};P) - V_l({\pi^*};P) ={(V_l({\pi_k};P) - V_{\bar l}({\pi_k};P_k))} + {(V_{\bar l}({\pi_k};P_k) - V_l({\pi^*};P))},$ where $V_{\bar l}$ is the value under a pessimistic bonus-augmented loss $\bar{l}_{k} = l_k - m T {\beta}^p_{k}$. We show that setting the scaling factor $m = T/(b_k-b_{\text{base}})$ is sufficient to guarantee $V_{\bar l}({\pi_k};P_k) \le V_l({\pi^*};P)$. In addition, the first error term $V_l({\pi_k};P) - V_{\bar l}({\pi_k};P_k)$ in the decomposition can be bounded by relating $V_{\bar l}$ back to $V_l$ via the classic value difference lemma~\cite{roknilamouki2025provably,shi2023nearsafe,sutton1998reinforcement} and the cumulative bonus. Finally, summing over $k$ and bounding the cumulative bonus terms $\sum \epsilon_k^{\pi_k}$ by invoking the pigeon-hole principle yields the final regret bound of $\tilde{\mathcal O}\left( T^3 S \sqrt{A K} / \gamma \right)$.
\end{proof}
\subsection{Upper-Level Regret Bound}\label{subsec:UL_bound}

We now bound the upper-level OCO provisioning regret.

\begin{lemma}[Upper-level BLOL Regret with Switching Costs]\label{lem:oco_regret}
Surrogate effective cost $\hat g_k(\cdot)$ is $\theta_g$-strongly convex on $\mathcal{B}$ and has bounded subgradients $\|\partial \hat g_k(b)\|\le G$ for all $b\in\mathcal{B}$ and all $k$. Let $\{b_k\}$ be updated by projected subgradient descent using the approximate subgradient $\hat h_k$ and step-size $\eta_k=\frac{1}{\theta_g(k-K_0)}$ for $k>K_0$. Then, with probability at least $1-3\delta$, we have
\begin{align}
\regr_{\mathrm{UL}}(K,b) \le \underbrace{G^2 \log K / (2\theta_g) + M}_{\text{Surrogate regret and switching costs}} + \underbrace{\tilde{\mathcal O} \left(ST^{3}\sqrt{A K} / \gamma\right)}_{\text{Value approximation error}}, \nonumber
\end{align}
where $M = K_0 G(T-B_{0}) + G^2 \pi^2 / (6 \theta_g^2)$ collects constant overheads from the warm-up phase and the finite switching cost series, and $\pi$ is the mathematical constant.
\end{lemma}

Lemma~\ref{lem:oco_regret} isolates two phenomena. First, the term for surrogate regret and switching costs shows that, despite switching costs, the upper-level optimization error scales only as $\tilde{\gO}(\log K)$ under strong convexity. The key reason switching costs do not destroy the rate is stability, i.e., with $\eta_k\propto 1/k$, budget updates shrink over time, so the path length $\sum_k\|b_k-b_{k-1}\|^2$ is bounded by a convergent series. Second, the term for value approximation error captures the penalty for optimizing the surrogate $\hat g_k$ instead of the true $g_k$. This term scales as $\tilde{\mathcal O}(\sqrt{K})$ and matches the lower-level learning regret, which confirms that the overall performance is dominated by the difficulty of learning dynamics and the resulting sensitivity. Please see Appendix~\ref{F} for the complete proof of Lemma~\ref{lem:oco_regret}.

\begin{proof}[Proof Sketch of Lemma~\ref{lem:oco_regret}]
The core idea is to decompose the regret against the true objective $g_k$ into the regret on the surrogate $\hat g_k$ and the approximation error. Specifically, we have
\begin{align}
\regr_{\mathrm{UL}}(K,b) & = \underbrace{\sum\nolimits_{k} (\hat g_k(b_k) - \hat g_k(b^*)) + \text{Switching costs}}_{\text{Surrogate regret and switching costs}} \nonumber \\
&+ \underbrace{\sum\nolimits_{k} (g_k(b_k) - \hat g_k(b_k)) - (g_k(b^*) - \hat g_k(b^*))}_{\text{Value approximation error}}. \nonumber
\end{align}

\subsubsection{Surrogate regret and switching costs} Since $\hat h_k$ is a valid subgradient of the strongly convex surrogate $\hat g_k$ (as discussed in Sec.~\ref{subsec:upper_oco}), the first term can bounded as follows. With step size $\eta_k \propto 1/k$, the surrogate regret scales as $\log K$. The switching cost at step $k$ is bounded by $\|b_k - b_{k-1}\|^2 \le (G \eta_{k-1})^2 \propto 1/k^2$. Since the series $\sum k^{-2}$ converges, the cumulative switching cost is bounded by a constant.

\subsubsection{Value approximation error} The remaining terms capture the value difference between the true and surrogate objectives. By definition, $g_k(b) - \hat g_k(b) = \beta(L_k^*(b) - \hat L_k(b))$. Using the optimism lemma from the lower-level analysis (Please see Appendix \ref{thm:bdope_main_1} for complete proof of Lemma~\ref{lem:optimism}), we can show that $\hat L_k(b)$ (the value of the optimistic surrogate) is close to $L_k^*(b)$ (the true optimal value). The cumulative sum of these differences/errors is bounded by the lower-level regret bound, which thus yields the $\tilde{\mathcal{O}}(\sqrt{K})$ term in the regret.
\end{proof}

\subsection{Final Regret Bound and Safety Guarantee}\label{subsec:total_bound}

Combining (\ref{eq:regret_decomp})-(\ref{eq:oracle_cost_identity}), Lemma~\ref{thm:bdope_main}, and Lemma~\ref{lem:oco_regret}, we obtain the final performance and safety guarantee.

\begin{theorem}[Total regret of our bi-level algorithm BLOL]\label{thm:total_regret}
With probability at least $1-3\delta$, the regret of the proposed bi-level algorithm BLOL (Algorithm~\ref{alg:upper_level_framework}) satisfies
\begin{align}
\regr(K,BLOL) \le \tilde{\mathcal O} \left(S T^{3}\sqrt{A K} / \gamma \right) + \tilde{\gO}(\log K). \label{eq:total_regret_bound}
\end{align}
\end{theorem}

Theorem~\ref{thm:total_regret} establishes a near-optimal $\tilde{\mathcal O}(\sqrt{K})$ regret, implying the average regret vanishes. Importantly, the bound separates the roles of the two levels: the upper level contributes only logarithmic regret (and bounded switching overhead) thanks to strong convexity and stable updates, while the dominant $\tilde{\mathcal O}(\sqrt{K})$ term comes from learning the unknown lower-level dynamics safely even under time-varying budgets. This clean separation clarifies the benefit of our primal-dual interface. The extended LP provides a dual variable signal that converts lower-level learning progress into correct sensitivity feedback for upper-level subgradient in budget updating. Please see Appendix~\ref{total_regret_1} for the complete proof of Theorem~\ref{thm:bdope_main}.

\begin{theorem}[Safety Guarantee] \label{safety_guarantee}
With probability at least $1-3\delta$, the algorithm BALDE (Algorithm~\ref{alg:bdope_subroutine}) satisfies the budget constraint for all episodes $k \in [K]$, i.e., $V_d(\text{BALDE};P) \le b_k$.
\end{theorem}

Theorem~\ref{safety_guarantee} establishes the safety of our BALDE subroutine under dynamic/time-varying budget and constraint threshold. It guarantees that the lower-level scheduler strictly adheres to the resource limits imposed by the upper level. Please see Appendix~\ref{safety_guarantee_1} for the complete proof of Theorem~\ref{safety_guarantee}.

\begin{proof}[Proof sketch of Theorem~\ref{safety_guarantee}]
For the warm-up phase ($k \le K_0$), safety is guaranteed by baseline policy $\pi_{\mathrm{base}}$. For $k >K_0$, applying the value difference lemma and Hölder's inequality, the difference between the cost under true transition kernel $P$ and that under $P_k$ can be bound as $\left| V_d({\pi_k};P) - V_d({\pi_k};P_k) \right| \le \mathbb{E}[\sum_{t=1}^{T} \|P_t - P_{k,t}\|_1 \| V_{d,t+1}^{\pi_k} \|_\infty ] \le T \mathbb{E}_{\pi_k, P_k}[ \sum_{t=1}^{T} \bar{\beta}_{t,k}^{p}] = \epsilon_k^{\pi_k}(P_k)$, where $V_{d,t}^{\pi} \triangleq \E_{\pi,P}[ \sum\nolimits_{\tau=t}^T d_t(s_t,a_t) ]$. Thus, we can always get $V_d({\pi_k};P)\le V_d({\pi_k};P_k)+\epsilon_k^{\pi_k}(P_k)=V_{\bar{d}}(\pi_k;P_k) \le b_k$. Hence, the budget constraint is satisfied for all episodes with high probability.
\end{proof}

\section{Numerical Results}

\begin{figure*}[t] 
    \centering
 \includegraphics[width=1\textwidth]{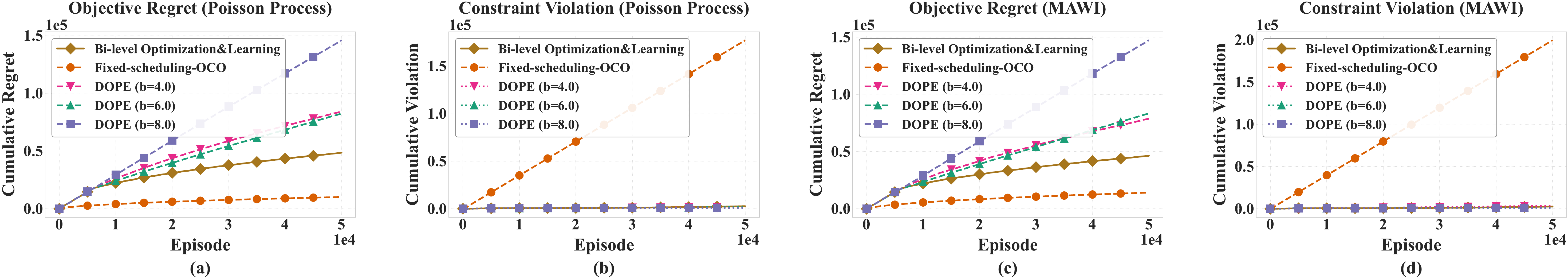} 
    \caption{Cumulative objective gap and cumulative budget violation under (a,b) Poisson arrivals and (c,d) MAWI traces. Here,  $\rho_k=5.0+0.5\sin(2\pi k/2000)+\epsilon_k$,  $\epsilon_k\sim\mathcal{N}(0,0.1^2)$, $M_0=0.25$, $M_1=0.01$, $M_2=0.05$, $\alpha=0.5$, $\beta=1.0$. }

    \label{fig:results} 
\end{figure*}

We evaluate the proposed bi-level algorithm (Algorithm~\ref{alg:upper_level_framework}) on a queueing-based slice scheduling model under both Poisson arrivals (with mean $\lambda=1.12$ packets/slot) and real traffic traces MAWI~\cite{traffictrace} (aggregated into $10$ms bins and linearly scaled to mean $1.12$ while preserving burstiness). The goal is to jointly provision an episode-level bandwidth budget with reconfiguration overhead and schedule packet service within each episode under an episode-wise budget constraint.

\subsection{Experimental Configurations}

\subsubsection{Lower level (queue scheduling learning)} The state $s_t\in\{0,1,\ldots,S_{\max}\}$ is the queue backlog with $S_{\max}=10$. The action $a_t\in\{0,1,2\}$ is the number of RBs allocated in slot $t$. The queue evolves as $s_{t+1}=\min\{S_{\max},[s_t-a_t+A_t]^+\}$, where $[x]^+=\max\{x,0\}$ and $A_t$ is the packet arrival. We use the latency/backlog stage loss $l(s_t,a_t)=\mu+(1-\mu)(s_t/S_{\max})^2$ and the per-slot resource consumption $d(s_t,a_t)=a_t/2$, where $\mu=0.1$. We let the horizon $T=10$.

\subsubsection{Upper level (budget provisioning optimization)} At the beginning of episode $k$, the budget $b_k\in[B_{0},T]$ is provisioned. The cost $f_k(b)=M_1 b + M_2(b-\rho_k)^2 + M_0(b-\rho_0)^2$ accounts for linear resource expenditures ($M_1$) and Service Level Agreement tracking ($M_2, M_0$), where sinusoidal $\rho_k$ emulates diurnal seasonality for non-stationary evaluation. Across episodes we charge the switching cost $\alpha\|b_k-b_{k-1}\|^2$.

\subsection{Evaluation Metrics and Baselines}

We report the cumulative \emph{objective gap} relative to a optimal static solution and cumulative budget violation $\mathrm{Viol}(K)=\sum_{k=1}^K [V_d(\pi_k;P)-b_k]_+$ for all methods. We compare our algorithm with two baselines, fixed-budget DOPE~\cite{bura2022dope} with $b\in\{4,6,8\}$ (directly using DOPE with given fixed budget $b$ and without adaptive upper-level provisioning) and fixed-scheduling OCO (directly optimizing $f_k(\cdot)+\alpha\|b_k-b_{k-1}\|^2$ via OCO, while the lower level runs standard RL with $Q$-learning~\cite{sutton1998reinforcement} and without dual feedback coupling.

\subsection{Results}

We run $K=5\times 10^4$ episodes. Fig.~\ref{fig:results} reports the cumulative objective gap (a,c) and cumulative budget violation (b,d) under Poisson arrivals and MAWI traces. In both traffic settings, our bi-level optimization-learning method achieves the \emph{smallest objective gap among all budget-feasible algorithms} while keeping the cumulative violation essentially zero, indicating that dual-coupled provisioning can reduce long-run cost without sacrificing safety. In contrast, fixed-budget DOPE remains safe but incurs a substantially larger objective gap because the provisioned budget is not adapted across episodes. Finally, the fixed-scheduling OCO baseline attains a smaller objective gap by directly optimizing provisioning costs, but it persistently violates the episode budget, demonstrating that ignoring the cross-level coupling and lower-level safety/sensitivity feedback can lead to systematically infeasible provisioning under stateful queue dynamics and switching overhead.

\section{Conclusion and Future Work}

We studied a bi-level online provisioning and scheduling problem with switching costs and cross-level budget constraints. The proposed framework integrates slow-time-scale resource provisioning (modeled as OCO with switching penalties) with fast-time-scale, state-dependent scheduling under \emph{varying} constraint thresholds (modeled as CMDP under unknown dynamics). Our primal-dual bi-level algorithm couples the two levels through a dual sensitivity signal (the budget multiplier) returned by the lower level, enabling stable budget updates while maintaining safety during exploration. We established near-optimal static regret and high-probability satisfaction of the episode-wise budget constraints.

Future work includes the following directions. First, while the $\tilde{\mathcal O}(\sqrt{K})$ dependence is order-optimal, tightening the dependence on other problem parameters (e.g., $T,S,A$) remains an important objective. Second, we will extend the analysis to dynamic benchmarks to better capture non-stationary traffic and network conditions. Third, motivated by network slicing deployments, we plan to study multi-slice and distributed settings where provisioning decisions are coupled across slices.

\bibliographystyle{IEEEtran}
\bibliography{reference}

\input{Appendix}

\end{document}

%% file: mathcommand.tex
\usepackage{amsmath,amsfonts,bm}

\def\eqref#1{equation~\ref{#1}}

\def\1{\bm{1}}

\DeclareMathAlphabet{\mathsfit}{\encodingdefault}{\sfdefault}{m}{sl}
\SetMathAlphabet{\mathsfit}{bold}{\encodingdefault}{\sfdefault}{bx}{n}

\def\gB{{\mathcal{B}}}

\def\gO{{\mathcal{O}}}
\def\gP{{\mathcal{P}}}

\def\gW{{\mathcal{W}}}

\def\sA{{\mathbb{A}}}

\def\sS{{\mathbb{S}}}

\newcommand{\E}{\mathbb{E}}

\DeclareMathOperator*{\argmin}{arg\,min}

%% file: Appendix.tex
\appendices

\section{Proof of Lemma~\ref{thm:bdope_main}}\label{thm:bdope_main_1}
 
\subsection{\textbf{High probability events}}\label{high_probability_event}
In this subsection, we define a high-probability event $\mathcal{G}$, which ensures that our estimations of the transition dynamics and state-action visitations are highly accurate. Conditioned on $\mathcal{G}$, we show that the regret of the lower-level algorithm is bounded.

To construct variance-aware confidence intervals for the unknown transition dynamics $P$, we bound the estimation error of the empirical model using the empirical Bernstein inequality.
\begin{lemma}[Empirical Bernstein Inequality~\cite{maurer2009empirical}]\label{lem:bernstein}
Let $X=(X_1,\ldots,X_n)$ be an i.i.d. sequence of random variables with values in $[0,1]$,
and let $\delta\in(0,1)$.
Then, with probability at least $1-\delta$:
\begin{align}
\Big|\,\mathbb{E}[X]-\frac{1}{n}\sum_{i=1}^{n}X_i\,\Big|
\le
\sqrt{\frac{2V_n(X)\log(2/\delta)}{n}} + \frac{7\log(2/\delta)}{3(n-1)}.\nonumber
\end{align}
where $V_n(X)$ denotes the empirical variance of $X$.
\end{lemma}

Building upon Lemma~\ref{lem:bernstein}, we construct the confidence set $\mathcal{P}_k$ centered
around the empirical transition model $\hat P_{k,t}$ as:
$\mathcal{P}_k
= \bigcap_{t\in[T],(s,a)\in\sS\times\sA} \mathcal{P}_{k,t}(s,a),$ where the component set is defined as:
\begin{align}\label{eq:Pkk}
\mathcal{P}_{k,t}(s,a)
= \big\{P' : |P'_{t}(s'|s,a)-\hat P_{k,t}(s'|s,a)|
\le \beta^{p}_{k,t}(s,a,s')\Big\} \nonumber
\end{align}
The confidence radius $\beta^{p}_{k,t}(s,a,s')$ is given by:
\begin{align}
\beta^{p}_{k,t}(s,a,s')
&= \sqrt{\frac{4\,\mathrm{Var}(\hat P_{k,t}(s'|s,a))\,L'}
{n_{k,t}(s,a)\vee 1}}
+ \frac{14L'}{3\,(n_{k,t}(s,a)\vee 1)}
\nonumber
\end{align}
where $L'=\log\Big(\frac{2SA T K}{\delta}\Big),$ and the empirical variance is $\mathrm{Var}(\hat P_{k,t}(s'|s,a))=\hat P_{k,t}(s'|s,a)\big(1-\hat P_{k,t}(s'|s,a)\big).$ 

First, we define the event $F^p$ where the true dynamics fall within the confidence intervals:
\begin{align}
F^p = \big\{ P\in\mathcal{P}_k,~\forall k\in[K]\big\}.\nonumber
\end{align}
By invoking Lemma~\ref{lem:bernstein} and applying a union bound, we have $\Pr(F^p)\ge 1-2\delta$.

Next, we address the concentration of state-action visitation counters. We define the event $F^w$ regarding the visitation frequency
of each $(s,a)$ pair:
\begin{align}
F^w
=
\Bigg\{
n_{k,t}(s,a)
\ge
\frac{1}{2}\sum_{j<k} w_{j,t}(s,a)
- T\log\!\Big(\frac{SAT}{\delta}\Big)
\Bigg\}.\nonumber
\end{align}
where $\forall(t,s,a,k)\in[T]\times\sS\times\sA\times[K]$, $w_{j,t}(s,a)$ denotes the occupancy measure induced by the policy in episode $j$.

The following lemma establishes that $F^w$ holds with high probability. Its proof directly follows from \cite[Corollary E.4.]{dann2017unifying}.
\begin{lemma}\label{lem:occupancy_event}
Let $F^w$ be defined as above.
Then, with probability at least $1-\delta$, the event holds, i.e., $\Pr(F^w)\ge 1-\delta$.
\end{lemma}
Finally, we define the \emph{good event} as the intersection of the $F^p$ and $F^w$:
\begin{align*}
\mathcal G = F^p \cap F^w.    
\end{align*}

Since $\Pr(F^p)\ge 1-2\delta$ and $\Pr(F^w)\ge 1-\delta$, a standard union bound over $F^p$ and $F^w$ yields the yields the result formally stated in the following lemma.

\begin{lemma}[Good Event]\label{lem:good_event}
Let $\mathcal{G}$ be defined as above.
Then, with probability at least $1-3\delta$, the event
$\mathcal G=F^p\cap F^w$ holds, i.e., $\Pr(\mathcal G)\ge 1-3\delta.$

\end{lemma}

Consequently, all subsequent theoretical analysis and regret bounds for the BALDE algorithm are derived conditioned on this high-probability event $\mathcal G$.

\subsection{\textbf{Auxiliary Lemmas}} In this subsection, we present several technical lemmas that serve as the building blocks for our regret analysis.
\begin{lemma}[Lemma 36, \cite{efroni2020exploration}]\label{lem:sum_sqrt_n}
Conditioned on the event $F^w$, the visitation frequencies satisfy:
\begin{align}
\sum_{k=1}^{K}\sum_{t=1}^{T}
\mathbb{E}\!\left[\frac{1}{\sqrt{n_{k,t}(s_{k,t},a_{k,t})\vee1}}
\right]
\le
\tilde{\mathcal O}\!\big(\sqrt{SAT^2K}+SAT\big).\nonumber
\end{align}
\end{lemma}

\begin{lemma}[Lemma 37, \cite{efroni2020exploration}]\label{lem:sum_inv_n}
Conditioned on the event $F^w$, the following bound also holds:
\begin{align}
\sum_{k=1}^{K}\sum_{t=1}^{T}
\mathbb{E}\!\left[\frac{1}{n_{k,t}(s_{k,t},a_{k,t})\vee1}
\right]
\le
\tilde{\mathcal O}(SAT^2).\nonumber
\end{align}
\end{lemma}

\begin{lemma}[Value Difference Lemma]\label{lem:value_difference}
Consider two MDPs
$M=(\sS,\sA,l,P)$ and $M'=(\sS,\sA,l,P')$ sharing the same state-action space and loss function $l$, but with different transition dynamics $P$ and $P'$. For any policy $\pi$, any state $s\in\sS$, and any time step $t\in[T]$, let $V_{l,t}^{\pi}(s;P) \triangleq \mathbb{E}_{\pi,P}[\sum_{\tau=t}^T l_\tau(s_\tau,a_\tau) \mid s_t=s]$ denote the value function from step $t$. Then, the following equality holds:
\begin{align}
&V^{\pi}_{l,t}(s;P) - V^{\pi}_{l,t}(s;P')
\nonumber\\&=
\mathbb{E}_{\pi,P'}\!\left[
\sum_{\tau=t}^{T}
\big((P_{\tau}-P'_{\tau})(\cdot|s_{\tau},a_{\tau})\big)^{\!\top}
V^{\pi}_{l,\tau+1}(\cdot;P)
~\Big|~s_t=s\right]\nonumber \\
&=
\mathbb{E}_{\pi,P}\!\left[
\sum_{\tau=t}^{T}
\big((P'_{\tau}-P_{\tau})(\cdot|s_{\tau},a_{\tau})\big)^{\!\top}
V^{\pi}_{l,\tau+1}(\cdot;P')
~\Big|~s_t=s\right].\nonumber
\end{align}
\end{lemma}
The proof of lemma \ref{lem:value_difference} is below:
\begin{proof}
Fix a policy $\pi$, a state $s \in \mathcal{S}$, and a time step $t \in [T]$. To simplify notation, we define:
\[
V_t(s; P) \triangleq V_{l,t}^{\pi}(s;P), \quad V_t(s; P') \triangleq V_{l,t}^{\pi}(s;P').
\]
and the value difference \[\Delta_t(s) \triangleq V_t(s; P) - V_t(s; P').\]
We use the finite-horizon convention $V_{T+1}(\cdot;P)=V_{T+1}(\cdot;P')\equiv 0$.

By definition, the value function satisfies the Bellman equation. For model $P$, expanding the expectation over next states yields:
\begin{align}
V_t(s; P) &= \mathbb{E}_{\pi} \left[ l_t(s,a) + \sum_{s' \in \sS} P_t(s'|s,a) V_{t+1}(s'; P) \,\middle|\, s_t = s \right] \nonumber \\
&= \mathbb{E}_{\pi} \left[ l_t(s,a) + \big( P_t(\cdot|s,a) \big)^\top V_{t+1}(\cdot; P) \,\middle|\, s_t = s \right]. \label{eq:bellman_P}
\end{align}
Similarly, for model $P'$, the Bellman equation is:
\begin{align}
V_t(s; P') &= \mathbb{E}_{\pi} \left[ l_t(s,a) + \big( P'_t(\cdot|s,a) \big)^\top V_{t+1}(\cdot; P') | s_t = s \right]. \label{eq:bellman_P_prime}
\end{align}
Here, $\E_{\pi}[\cdot\mid s_t=s]$ denotes expectation w.r.t. $a\sim \pi_t(\cdot\mid s)$ only;
the next-state expectation under $P_t(\cdot\mid s,a)$ (resp. $P'_t(\cdot\mid s,a)$) is already taken
inside the inner product $\big( P_t(\cdot|s,a) \big)^\top V_{t+1}(\cdot; P) $ (resp. $\big( P'_t(\cdot|s,a) \big)^\top V_{t+1}(\cdot; P') $).

Subtracting (\ref{eq:bellman_P_prime}) from (\ref{eq:bellman_P}), and noting that the $l_t$ terms cancel, we obtain:
\begin{align} \label{difference}
\Delta_t(s) &= \mathbb{E}_{\pi} \bigg[ \big( P_t(\cdot|s,a) \big)^\top V_{t+1}(\cdot; P) \nonumber \\
&\quad - \big( P'_t(\cdot|s,a) \big)^\top V_{t+1}(\cdot; P') \,\bigg|\, s_t = s \bigg].
\end{align}
Next, we add and subtract the term $\big( P'_t(\cdot|s,a) \big)^\top V_{t+1}(\cdot; P)$ inside the expectation:
\begin{align}
\Delta_t(s) &= \mathbb{E}_{\pi} \Biggl[
    \underbrace{\left( \big( P_t(\cdot|s,a) \big)^\top - \big( P'_t(\cdot|s,a) \big)^\top \right) V_{t+1}(\cdot; P)}_{\text{(a)}} \nonumber \\
    &\quad + \underbrace{\big( P'_t(\cdot|s,a) \big)^\top \left( V_{t+1}(\cdot; P) - V_{t+1}(\cdot; P') \right)}_{\text{(b)}}
| s_t=s\Biggr]. \label{eq:decomp_step}
\end{align}

Let us analyze term (b) in (\ref{eq:decomp_step}). Note that the term $\big( P'_t(\cdot|s,a) \big)^\top (V_{t+1}(\cdot; P) - V_{t+1}(\cdot; P'))$ is exactly the expected value of the next-step difference $\Delta_{t+1}(s_{t+1})$, where the next state is sampled from $P'$. Specifically:
\[
\text{(b)} = \mathbb{E}_{ s' \sim P'_t(\cdot|s,a)} [ \Delta_{t+1}(s') ].
\]
Substituting this back into (\ref{eq:decomp_step}), we can rewrite the difference as an expectation over the trajectory generated by $P'$:
\begin{align}
\Delta_t(s) = & \mathbb{E}_{\pi, P'} \Biggr[
    \big( (P_t - P'_t)(\cdot|s_t,a_t) \big)^\top V_{t+1}(\cdot; P)\nonumber\\
    & + \Delta_{t+1}(s_{t+1})
| s_t = s \Biggr]. \label{eq:recursion}
\end{align}

Equation (\ref{eq:recursion}) establishes a recursive relation for $\Delta_t(s)$. Since $V_{T+1} \equiv 0$ implies $\Delta_{T+1} \equiv 0$, we can unroll this recurrence forward in time from $t$ to $T$:
\begin{align}
\Delta_t(s) &= \mathbb{E}_{\pi, P'} \left[ \sum_{\tau=t}^T \big( (P_\tau - P'_\tau)(\cdot|s_\tau,a_\tau) \big)^\top V_{\tau+1}(\cdot; P) \,\middle|\, s_t = s \right] \nonumber
\end{align}
This proves the first equality of the lemma. The second equality follows from a symmetric argument by adding and subtracting $\big( P_t(\cdot|s,a) \big)^\top V_{t+1}(\cdot; P')$ in (\ref{difference}).
\end{proof}

We define the cumulative transition bonus $\epsilon_k^{\pi}(P)$ for an episode $k$ under policy $\pi$ and model $P$ with initial state $s_1$ as:
\[
 \epsilon_k^{\pi}(P) \triangleq T\,\sum_{t=1}^T\mathbb{E}_{\pi, P}\!\bigg[ \sum_{s' \in \sS} \beta^{p}_{k,t}(s_{k,t},a_{k,t},s')
           \,\bigg|\,s_1\bigg].
\]
We now derive a high-probability upper bound for the cumulative term $\epsilon_k^{\pi}(P)$.
 
\begin{lemma}\label{lem:sum-eps}
Consider the sequence of policies $\{\pi_k\}_{k=1}^{K}$ generated by BALDE. For any $K'\le K$, with probability at least $1-3\delta$, the following bound holds:
\[
 \sum_{k=1}^{K'} \epsilon_k^{\pi_k}(P)
 \le \tilde{\mathcal O}\!\Big(S\sqrt{A\,T^{4}K'}\Big).
\]
\end{lemma}

\begin{proof}
Recall the definition of the confidence radius $\beta^{p}_{k,t}(s,a,s')$:
\begin{align}
\beta^{p}_{k,t}(s,a,s')
&= \sqrt{\frac{4\,\mathrm{Var}(\hat P_{k,t}(s'|s,a))\,L'}
{n_{k,t}(s,a)\vee 1}}
+ \frac{14L'}{3\,(n_{k,t}(s,a)\vee 1)}. \nonumber
\end{align}
Using the property $\mathrm{Var}(X) \le \mathbb{E}[X]$ for Bernoulli variables, we have $\mathrm{Var}(\hat P_{k,t}(s'|s,a)) \le \hat P_{k,t}(s'|s,a)$. Summing $\beta^{p}_{k,t}$ over all next states $s'$ yields:
\begin{align}
&\sum_{s'} \beta^{p}_{k,t}(s,a,s')
\le \sum_{s'} \sqrt{\frac{4 L' \hat P_{k,t}(s'|s,a)}{n_{k,t}(s,a)\vee 1}}
+ \sum_{s'} \frac{14L'}{3\,(n_{k,t}(s,a)\vee 1)} \nonumber \\
&= \sqrt{\frac{4L'}{n_{k,t}(s,a)\vee 1}} \sum_{s'} \sqrt{\hat P_{k,t}(s'|s,a)}
+ \frac{14 S L'}{3\,(n_{k,t}(s,a)\vee 1)}. \nonumber
\end{align}
By the Cauchy-Schwarz inequality, $\sum_{s'} \sqrt{\hat P_{k,t}(s'|s,a)} \le \sqrt{S \sum_{s'} \hat P_{k,t}(s'|s,a)} = \sqrt{S}$. Substituting this back into the definition of $\epsilon_k^{\pi_k}(P)$, we have:
\begin{align}
\sum_{k=1}^{K'} \epsilon_k^{\pi_k}(P)
&= T\sum_{k=1}^{K'} \sum_{t=1}^T \,\mathbb{E}_{\pi_k, P}\!\left[\sum_{s'} \beta^{p}_{k,t}(s_{k,t},a_{k,t},s') \right] \nonumber \\
&\le
T\sqrt{S}
\sum_{k=1}^{K'} \sum_{t=1}^T
\mathbb{E}_{\pi_k, P}\!\left[
\sqrt{\frac{4L'}{n_{k,t}(s_{k,t},a_{k,t})\vee 1}}
\right]\nonumber\\
&\quad +
\frac{14 S L' T}{3}
\sum_{k=1}^{K'} \sum_{t=1}^T
\mathbb{E}_{\pi_k, P}\!\left[
\frac{1}{n_{k,t}(s_{k,t},a_{k,t})\vee 1}
\right].\nonumber
\end{align}
Conditioned on the event $F^w$, we invoke Lemma~\ref{lem:sum_sqrt_n} and Lemma~\ref{lem:sum_inv_n} to bound the summation terms:
\begin{align}
\sum_{k=1}^{K'} \epsilon_k^{\pi_k}(P)
&\le
T\sqrt{S} \cdot \tilde{\mathcal{O}}\left(\sqrt{S A T^2 K'}+SAT\right) + \tilde{\mathcal{O}}(S^2 A T^3) \nonumber \\
&= \tilde{\mathcal{O}}\left(S\sqrt{A T^4 K'}\right). \nonumber
\end{align}
This concludes the proof.
\end{proof}

\subsection{\textbf{Proof of the Feasibility of the Extended LP Problem}}
Before analyzing the regret bound , we first establish that the lower-level optimization in BALDE is well-defined at every episode. In the online learning phase ($k>K_0$), BALDE solves the extended LP problem (\ref{eq:dop_lower}) at each episode. Because the pessimistic constraint costs are heavily inflated by uncertainty bonuses early in learning, the LP might  be infeasible in early learning phase. Thus, we need to show that this optimization problem is feasible after the $K_0$ warm-up phase.
The proposition below shows that, on the good event $\mathcal G$, the baseline policy $\pi_{base}$ remains feasible under the pessimistic constraint budget after sufficiently many warm-up episodes. Consequently, the extended LP admits at least one feasible solution for every episode $k>K_0$.

\begin{proposition}[Feasibility of the Extended LP problem]\label{prop:dop-feasible}
Under the BALDE algorithm, let $B_0$ denote the initial budget used in the warm-up phase, and let
$\pi_{\mathrm{base}}$ be a safe baseline policy satisfying
$V_d(\pi_{\mathrm{base}};P)=b_{\mathrm{base}}$ with slack
$\gamma \triangleq B_0 - b_{\mathrm{base}} > 0$.
Suppose $\pi_k=\pi_{\mathrm{base}}$ for all $k\le K_0$. Then, with probability at least $1-3\delta$ ,
the pair $(\pi_{\mathrm{base}},P)$ is a feasible solution to the extended LP problem
(\ref{eq:dop_lower}) for all episodes $k>K_0$, where $K_0 = \tilde{\mathcal O}\!\left(\frac{S^2 A T^4}{\gamma^2}\right).$

\end{proposition}

\begin{proof}
Conditioned on the event $\mathcal{G}$, the model constraint $P \in \mathcal{P}_k$ is satisfied. 
Recall that the pessimistic constraint cost is \[\bar d_{k,t}(s,a) = d_t(s,a) + T\beta^{p}_{k,t}(s,a),\] 
where $\beta^{p}_{k,t}(s,a)=\sum_{s'\in\mathcal S}\beta^{p}_{k,t}(s,a,s')$. By linearity of expectation:
\begin{align}
V_{\bar d}({\pi};P)
&= \mathbb{E}_{\pi, P} \left[ \sum_{t=1}^T \left( d_t(s_t,a_t) + T\bar\beta^{p}_{k,t}(s_t,a_t) \right) \right] \nonumber \\
&= V_{d}({\pi};P) + \epsilon_k^{\pi}(P).
\label{eq:V_d_decomp}
\end{align}

To prove feasibility of the extended LP, it suffices to show that for any episode $k > K_0$, the baseline pair $(\pi_{\mathrm{base}}, P)$ satisfies the pessimistic budget constraint:
\[
V_{\bar d}(\pi_{\mathrm{base}}; P) \le b_k.
\]
Since $B_0$ is the lower bound of the budgets (i.e., $b_k \ge B_0$ for all $k$), a sufficient condition for feasibility is:
\[
V_{\bar d}(\pi_{\mathrm{base}}; P) \le B_0.
\]
Using equation (\ref{eq:V_d_decomp}) and the property $V_d(\pi_{\mathrm{base}}; P) = b_{\mathrm{base}}$, this condition is equivalent to:
\begin{align}
b_{\mathrm{base}} + \epsilon_k^{\pi_{\mathrm{base}}}(P) \le B_0 \iff \epsilon_k^{\pi_{\mathrm{base}}}(P) \le \gamma. \label{eq:feasibility_condition}
\end{align}

We bound the number of episodes required to satisfy (\ref{eq:feasibility_condition}) by contradiction. Suppose the condition is violated for the first $K'$ episodes, i.e., $\epsilon_k^{\pi_{\mathrm{base}}}(P) > \gamma$ for all $k \le K'$, while running $\pi_k = \pi_{\mathrm{base}}$. Summing this inequality over $k$ yields:
\[
K' \gamma < \sum_{k=1}^{K'} \epsilon_k^{\pi_{\mathrm{base}}}(P).
\]
Applying Lemma~\ref{lem:sum-eps} with the sequence of policies $\pi_k = \pi_{\mathrm{base}}$, we have:
\[
\sum_{k=1}^{K'} \epsilon_k^{\pi_{\mathrm{base}}}(P) \le \tilde{\mathcal O}\!\left(S \sqrt{A T^4 K'}\right).
\]
Combining these inequalities implies $K' \gamma < \tilde{\mathcal O}(S \sqrt{A T^4 K'})$, which simplifies to:
\[
\sqrt{K'} < \tilde{\mathcal O}\left(\frac{S \sqrt{A T^4}}{\gamma}\right) \implies K' < \tilde{\mathcal O}\left(\frac{S^2 A T^4}{\gamma^2}\right).
\]
This leads to a contradiction provided that $K' \ge \tilde{\mathcal O}(S^2 A T^4 / \gamma^2)$. Thus, we can select $K_0 = \tilde{\mathcal O}(S^2 A T^4 / \gamma^2)$ to ensure the condition is met.

Finally, since the algorithm executes a fixed policy $\pi_{\mathrm{base}}$ during the warm-up phase, the visitation counts $n_{k,t}$ are non-decreasing in $k$. Consequently, the confidence radius and the cumulative bonus $\epsilon_k^{\pi_{\mathrm{base}}}(P)$ are monotonically non-increasing. Therefore, the condition $\epsilon_k^{\pi_{\mathrm{base}}}(P) \le \gamma$ holds for all $k \ge K_0$. This completes the proof.
\end{proof}

\subsection{\textbf{Proof of the CMDP Regret Bounds}}

We first define the regret metric for the lower-level BALDE algorithm. Let $\pi_k^*$ denote the optimal policy for the true CMDP under the episode budget $b_k$, and let $\pi_k$ be the policy executed by BALDE algorithm in episode $k$. The cumulative regret after $K$ episodes is defined as: $R(K) \triangleq \sum_{k=1}^{K} \bigl( V_{l}(\pi_k; P)-V_{l}(\pi_k^*; P) \bigr)$.

The cumulative regret can be decomposed into two parts: the warm-up phase and the online learning phase.
\begin{align}
R(K)
&=\sum_{k=1}^{K}
\bigl( V_{l}(\pi_k; P)-V_{l}(\pi_k^*; P) \bigr) \nonumber\\
&=\sum_{k=1}^{K_0}
\bigl( V_{l}(\pi_k; P)-V_{l}(\pi_k^*; P) \bigr) \nonumber\\
&\quad + \sum_{k=K_0+1}^{K}
\bigl( V_{l}(\pi_k; P)-V_{l}(\pi_k^*; P) \bigr).
\label{eq:regret-summary}
\end{align}

\paragraph{Warm-up Phase ($k \le K_0$)}
During this phase, the algorithm executes the baseline policy $\pi_k=\pi_{\mathrm{base}}$. Since the cumulative cost is bounded by $T$ (i.e., $|V_{l}(\pi_k; P)-V_{l}(\pi^*_k; P)| \le T$), we can bound the regret using the warm-up length $K_0$ derived in Proposition~\ref{prop:dop-feasible}:
\begin{align} \label{eq:k0-bound}
\sum_{k=1}^{K_0}
\bigl( V_{l}(\pi_k; P)-V_{l}(\pi^*_k; P) \bigr)
\le T K_0
\le \tilde{\mathcal{O}}\!\left(\frac{S^2 A T^5}{\gamma^2}\right).
\end{align}
where $\gamma \triangleq B_0 - b_{\mathrm{base}}$ is the safety slack.

\paragraph{Online Phase ($k > K_0$)}
For the subsequent episodes, we decompose the episode-wise regret:
\begin{align}
&V_{l}(\pi_k; P)-V_{l}(\pi^*_k; P)\nonumber\\
&=
\underbrace{
V_{l}(\pi_k; P)-V_{\bar l}(\pi_k; P_k)
}_{\Delta_k^{(1)}}
+
\underbrace{
V_{\bar l}(\pi_k; P_k)-V_{l}(\pi^*_k; P)
}_{\Delta_k^{(2)}}.\nonumber
\end{align}
where $V_{\bar l}(\pi_k; P_k)$ denotes the optimistic value function constructed with the bonus. Summing over $k > K_0$, we obtain:
\begin{align} \label{eq:regret-decomp-online}
&\sum_{k=K_0+1}^{K}
\bigl( V_{l}(\pi_k; P)-V_{l}(\pi^*_k; P) \bigr)\nonumber\\
&=
\sum_{k=K_0+1}^{K} \Delta_k^{(1)}
+
\sum_{k=K_0+1}^{K} \Delta_k^{(2)}.
\end{align}

We now verify the optimism of the algorithm, which ensures that the second term is non-positive.

\begin{lemma}\label{lem:optimism}
Let $(\pi_k,P_k)$ be the optimal solution to the DOP problem~(\ref{eq:dop_lower}) at episode $k$. By definition:
\begin{align}
(\pi_k,P_k)
\in \arg\min_{\pi,\;P'\in\mathcal{P}_k}
\Bigl\{
V_{\bar l}(\pi; P')
\;\text{s.t.}\;
V_{\bar d}(\pi; P') \le b_k
\Bigr\}. \nonumber
\end{align}
This implies $V_{\bar l}(\pi_k; P_k) \;\le\; V_l(\pi_k^*; P)$, and hence
$\Delta_k^{(2)}
= V_{\bar l}(\pi_k; P_k)-V_{l}(\pi_k^*; P) \;\le\; 0$, provided that the bonus scaling parameter $m$ is sufficiently large.
\end{lemma}

\begin{proof}
First, we define the bonus-augmented loss (used only in the proof) by
\begin{align}
&\tilde l_{k,t}(s,a) \;=\; l_{k,t}(s,a)\;-\; mT\,\beta^p_{k,t}(s,a),\nonumber\\
&\beta^p_{k,t}(s,a) \;=\; \sum_{s'} \beta^p_{k,t}(s,a,s').\nonumber
\end{align}
The optimal solution $(\tilde{\pi}_k,\tilde{P}_k)$ can be formulated as:
\begin{align} \label{eq:proof}
(\tilde{\pi}_k,\tilde{P}_k)
  &= \argmin_{\pi',\,P' \in \mathcal{P}_k}
      V_{\tilde{r}_k}({\pi'};P')
      \quad \text{subject to } 
      V_{\bar{b}_k}({\pi'};P') \le b_k.
\end{align}
We show that if $m \ge \frac{T}{b_k-b_{\mathrm{base}}}$, then $V_{\tilde l}(\tilde\pi_k; \tilde P_k) \le V_{l}(\pi_k^*; P)$.

Let $w^{\pi}(P)$ denote the occupancy measure of policy $\pi$ in model $P$. Specifically, let $w^{\pi_{\mathrm{base}}}(\cdot;P)$ and $w^{\pi_k^*}(\cdot;P)$ denote the occupancy measures of the baseline policy $\pi_{\mathrm{base}}$ and the optimal policy $\pi_k^*$, respectively.
For episode $k$, consider a scalar $\alpha_k \in [0,1]$ and construct the mixed occupancy measure:
\begin{align}
\tilde w_t(s,a) \;=\; (1-\alpha_k)\,w^{\pi_k^*}_t(s,a;P)\;+\;\alpha_k\,w^{\pi_{\mathrm{base}}}_t(s,a;P).\nonumber
\end{align}
and let $\tilde\pi$ be the policy induced by $\tilde w$. Since the value function is linear in the occupancy measure, we have:
\begin{align}
& V_{\bar d}(\tilde\pi; P)
=(1-\alpha_k)V_{\bar d}(\pi_k^*; P)+\alpha_k V_{\bar d}(\pi_{\mathrm{base}}; P) \nonumber\\
&=(1-\alpha_k)\big(V_{d}(\pi_k^*; P)+ \epsilon_k^{\pi_k^*}\big)+\alpha_k \big(V_{d}(\pi_{\mathrm{base}}; P)+\epsilon_k^{\pi_{\mathrm{base}}} \big)\nonumber\\
&\le (1-\alpha_k)\!\left(b_k+\epsilon_k^{\pi_k^*}\right)
   +\alpha_k\!\left(b_{\mathrm{base}}+\epsilon_k^{\pi_{\mathrm{base}}}\right) \nonumber\\
&= b_k +\epsilon_k^{\pi_k^*} -\alpha_k\epsilon_k^{\pi_k^*}+\alpha_k\!\left(\epsilon_k^{\pi_{\mathrm{base}}}-(b_k-b_{\mathrm{base}})\right).\nonumber
\end{align}
From Proposition~\ref{prop:dop-feasible}, when $k\ge K_0$, we have $\epsilon_k^{\pi_{\mathrm{base}}} \le B_0 - b_{\mathrm{base}}=\gamma$. Since $b_k \ge B_0$, it follows that $\epsilon_k^{\pi_{\mathrm{base}}}-(b_k-b_{\mathrm{base}})\le 0$. Therefore, if
\begin{align}
\alpha_k \;\ge\;
\frac{\epsilon_k^{\pi_k^*}}
{b_k-b_{\mathrm{base}}-\epsilon_k^{\pi_{\mathrm{base}}}+\epsilon_k^{\pi_k^*}}.\nonumber
\end{align}
ensures $V_{\bar d}(\tilde\pi; P)\le b_k$. Thus, $(\tilde\pi,P)$ is a feasible solution for the  problem~(\ref{eq:proof}) at episode $k$.

Since $(\tilde\pi_k, \tilde P_k)$ is the optimal solution minimizing $V_{\tilde l}$, we have $V_{\tilde l}(\tilde\pi_k; \tilde P_k) \le V_{\tilde l}(\tilde{\pi}_k; P)$. Thus, it suffices to find $m$ such that $V_{\tilde l}(\tilde{\pi_k}; P) \le V_{l}(\pi_k^*; P)$.
By linearity of the occupancy measure:
\begin{align}
V_{\tilde l}(\tilde\pi_k; P)
&=(1-\alpha_k)V_{\tilde l}(\pi_k^*; P)+\alpha_k V_{\tilde l}(\pi_{\mathrm{base}}; P)\nonumber \\
&= (1-\alpha_k)\!\left(V_{l}(\pi_k^*; P)-m\,\epsilon_k^{\pi_k^*}\right)\nonumber\\
  &\quad+\alpha_k\!\left(V_{l}(\pi_{\mathrm{base}}; P)-m\,\epsilon_k^{\pi_{\mathrm{base}}}\right) \nonumber \\
&\le V_{l}(\pi_k^*; P).
\end{align}
This inequality holds if:
\begin{align}
    m(1-\alpha_k)\epsilon_k^{\pi_k^*}+m\alpha_k\epsilon_k^{\pi_{\mathrm{base}}} &\ge \alpha_k(V_{l}(\pi_{\mathrm{base}}; P)-V_{l}(\pi_k^*; P)). \nonumber
\end{align}
We just setting $\alpha_k \;=\;
\frac{\epsilon_k^{\pi_k^*}}
{b_k-b_{\mathrm{base}}-\epsilon_k^{\pi_{\mathrm{base}}}+\epsilon_k^{\pi_k^*}}$, solving for $m$:
\begin{align}
    m & \ge\frac{\alpha_k(V_{l}(\pi_{\mathrm{base}}; P)-V_{l}(\pi_k^*; P)) }{(1-\alpha_k)\epsilon_k^{\pi_k^*}+\alpha_k\epsilon_k^{\pi_{\mathrm{base}}}}\nonumber \\
    & =\frac{V_{l}(\pi_{\mathrm{base}}; P)-V_{l}(\pi_k^*; P) }{\frac{1-\alpha_k}{\alpha_k}\epsilon_k^{\pi_k^*}+\epsilon_k^{\pi_{\mathrm{base}}}}\nonumber \\
    & =\frac{V_{l}(\pi_{\mathrm{base}}; P)-V_{l}(\pi_k^*; P)}{b_k-b_{\mathrm{base}}}.
\end{align}
where the simplification uses the definition of $\alpha_k$.
Since $l \in [0,1]$, we have $V_{l}(\pi_{\mathrm{base}}; P) \le T$ and $V_{l}(\pi_k^*; P) \ge 0$. Thus, $m = \frac{T}{b_k-b_{\mathrm{base}}}$ satisfies the condition.
This concludes the proof of Lemma~\ref{lem:optimism}.
\end{proof}

Using Lemma~\ref{lem:optimism}, we can bound (\ref{eq:regret-decomp-online}) as follows:
\begin{align} \label{eq:delta}
\sum_{k=K_0+1}^{K}
\bigl( V_{l}({\pi_k};P)-V_{l}({\pi_k^*};P) \bigr)
\le
\sum_{k=K_0+1}^{K} \Delta_k^{(1)}.
\end{align}

Recall that $\Delta_k^{(1)} = V_{l}({\pi_k};P)-V_{\bar l}({\pi_k};P_k)$. Using Lemma~\ref{lem:sum-eps}, we have:
\begin{align} \label{tilde}
&\sum_{k=K_0+1}^{K} \Delta_k^{(1)} \nonumber\\
&= \sum_{k=K_0+1}^{K} \left(V_{l}({\pi_k};P)-V_{l}({\pi_k};P_k) + \frac{T}{b_k-b_{\mathrm{base}}}\epsilon_k^{\pi_k}(P_k) \right) \nonumber \\
& \le \sum_{k=K_0+1}^{K} \big(V_{l}({\pi_k};P)-V_{l}({\pi_k};P_k)\big) + \tilde{\mathcal{O}}\!\left(\frac{ST^3\sqrt{AK}}{\gamma} \right).
\end{align}
where $\gamma = B_0 - b_{\mathrm{base}}$ is the safety slack.

Next, we bound the estimation error term $\sum_{k=K_0+1}^{K} (V_{l}({\pi_k};P)-V_{l}({\pi_k};P_k))$.
Applying the Value Difference Lemma (Lemma~\ref{lem:value_difference}) and Hölder's inequality:
\begin{align} \label{final_bound}
 &\sum_{k=K_0+1}^{K} (V_{l}({\pi_k};P)-V_{l}({\pi_k};P_k))   \nonumber\\
 &=\sum_{k=K_0+1}^{K}\sum_{t=1}^{T}\mathbb{E}_{\pi_k,P}\!\left[
\big((P_{t}-P_{t,k})(\cdot|s_{t},a_{t})\big)^{\!\top}
V^{\pi_k}_{l,t+1}(\cdot;P_k)
\right]\nonumber \\
&\le \sum_{k=K_0+1}^{K}\sum_{t=1}^{T} \mathbb{E}_{\pi_k,P}\!\left[ \|P_t-P_{t,k}\|_1 \cdot \|V_{l,t+1}^{\pi_k}\|_\infty \right] \nonumber \\
&\le \sum_{k=K_0+1}^{K} \epsilon_k^{\pi_k}(P) \le \tilde{\mathcal{O}}\big(S T^2\sqrt{AK}\big).
\end{align}

Finally, combining Eq.~(\ref{eq:regret-summary}), Eq.~(\ref{eq:k0-bound}), Eq.~(\ref{eq:delta}), Eq.~(\ref{tilde}), and Eq.~(\ref{final_bound}), we obtain the total regret bound:
\begin{align}
R(K) &= \sum_{k=1}^{K}\!\Big(V_{l}({\pi_k};P)-V_l({\pi_k^*};P)\Big) \notag \\
&\le \tilde{\mathcal{O}}\!\left(\frac{S^2 A T^5}{\gamma^2}\right) +
\tilde{\mathcal{O}}\big(S T^2\sqrt{AK}\big)
+ \tilde{\mathcal{O}}\!\left(\frac{S T^3 \sqrt{AK}}{\gamma} \right) \notag \\
& \le \tilde{\mathcal{O}}\!\left(\frac{S T^3 \sqrt{AK}}{\gamma} \right).
\label{eq:model-gap-budget-final}
\end{align}
The bound (\ref{eq:model-gap-budget-final}) holds with probability at least $1-3\delta$.
\hfill$\square$

\section{Proof of Lemma~\ref{lem:oco_regret}} \label{F}
\subsection{\textbf{Convexity and Subgradients of the Lower-Level Value Function $L_k^*(b)$}}
\begin{lemma} \label{lem:Lk-convex}

$L_k^*(b)$ is convex for any $b > 0$.

\end{lemma}
\begin{proof}
    To prove convexity, we show that for any $b_1, b_2 > 0$ and any $\theta \in [0,1]$,
\[
L_k^*(\theta b_1 + (1-\theta)b_2) \le \theta L_k^*(b_1) + (1-\theta)L_k^*(b_2).
\]
Recall the definitions of expected loss and consumption:
\begin{align*}
J(\pi) &\triangleq \mathbb{E}_{\pi}\Big[\sum_{t=1}^T l_t(s_{k,t}, a_{k,t})\Big], \\
D(\pi) &\triangleq \mathbb{E}_{\pi}\Big[\sum_{t=1}^T d_t(s_{k,t}, a_{k,t})\Big].
\end{align*}
Let $\pi^{A,1}$ and $\pi^{A,2}$ be optimal policies under budgets $b_1$ and $b_2$, respectively. By definition, we have:
\begin{align*}
L_k^*(b_1) &= J(\pi^{A,1}), \quad D(\pi^{A,1}) \le b_1, \\
L_k^*(b_2) &= J(\pi^{A,2}), \quad D(\pi^{A,2}) \le b_2.
\end{align*}
Consider a mixed budget $\bar{b} = \theta b_1 + (1-\theta)b_2$. We construct a composite policy $\pi^{\bar{A}}$ as follows: before the episode starts, draw a Bernoulli random variable $U$ with $\Pr(U=1)=\theta$ and $\Pr(U=0)=1-\theta$. If $U=1$, execute $\pi^{A,1}$; otherwise, execute $\pi^{A,2}$.
By the law of total expectation, the resource consumption of $\pi^{\bar{A}}$ is:
\begin{align*}
D(\pi^{\bar{A}}) &= \theta D(\pi^{A,1}) + (1-\theta) D(\pi^{A,2}) \\
&\le \theta b_1 + (1-\theta) b_2 = \bar{b}.
\end{align*}
Thus, $\pi^{\bar{A}}$ is feasible for budget $\bar{b}$. Since $L_k^*(\bar{b})$ is the infimum over all feasible policies, we have $L_k^*(\bar{b}) \le J(\pi^{\bar{A}})$. Similarly, the expected loss is:
\begin{align*}
J(\pi^{\bar{A}}) &= \theta J(\pi^{A,1}) + (1-\theta) J(\pi^{A,2}) \\
&= \theta L_k^*(b_1) + (1-\theta) L_k^*(b_2).
\end{align*}
Combining these yields $L_k^*(\bar{b}) \le \theta L_k^*(b_1) + (1-\theta) L_k^*(b_2)$, which completes the proof of convexity.
\bigskip
\end{proof}

\begin{lemma}(Subgradient of the value function $L_k^*(b)$)
\label{lem:Lk-subgradient}
For each episode $k$, any optimal dual multiplier $\lambda^*(b)$ associated with
the budget constraint satisfies
$-\lambda^*(b)\in\partial L_k(b).$
\end{lemma}

\begin{proof}
For brevity, we omit the episode index $k$. The primal problem is defined as $L(b) = \inf_{\pi: D(\pi) \le b} J(\pi)$.
We introduce a Lagrange multiplier $\lambda \ge 0$ and define the Lagrangian:
\[
\mathcal{L}(\pi, \lambda; b) = J(\pi) + \lambda(D(\pi) - b).
\]
The dual function is $g(\lambda; b) = \inf_{\pi} \mathcal{L}(\pi, \lambda; b) = \psi(\lambda) - \lambda b$, where $\psi(\lambda) \triangleq \inf_{\pi} (J(\pi) + \lambda D(\pi))$.
Assuming Slater's condition holds, strong duality implies:
\begin{equation} \label{eq:strong_duality}
L(b) = \max_{\lambda \ge 0} (\psi(\lambda) - \lambda b).
\end{equation}
Let $\lambda^*(b)$ be an optimal dual multiplier for a fixed budget $b$. Then, $L(b) = \psi(\lambda^*(b)) - \lambda^*(b)b$.
Now, consider any arbitrary budget $\tilde{b} \in \mathcal{B}$. By (\ref{eq:strong_duality}), we have:
\begin{align*}
L(\tilde{b}) &= \max_{\lambda \ge 0} (\psi(\lambda) - \lambda \tilde{b}) \\
&\ge \psi(\lambda^*(b)) - \lambda^*(b) \tilde{b}.
\end{align*}
Substituting $\psi(\lambda^*(b)) = L(b) + \lambda^*(b)b$ into the inequality:
\begin{align*}
L(\tilde{b}) &\ge L(b) + \lambda^*(b)b - \lambda^*(b)\tilde{b} \\
&= L(b) + (-\lambda^*(b))(\tilde{b} - b).
\end{align*}
By the definition of the subgradient, this inequality implies that $-\lambda^*(b) \in \partial L(b)$.
Restoring the episode index, we have $-\lambda_k^*(b_k) \in \partial L_k^*(b_k)$.
\end{proof}

\subsection{\textbf{Proof of the OCO with Switching Cost Regret Bounds}}
\begin{proof}
It suffices to prove the regret bound of OCO with switching costs. Recall that the regret is given by:
\begin{align}
    \text{Reg}_{\text{UL}}(K)
  &= \sum_{k=1}^K
  \Big(\big(
    f_k(b_k)
    + \alpha\|b_k - b_{k-1}\|^2
    + \beta L_k^*(b_k)\big)
    \nonumber\\
    &\quad
    - \big(f_k(b^*)
    + \beta L_k^*(b^*)
  \big)\Big)
  \nonumber \\
  &=\sum_{k=1}^{K}(g_k(b_k)-g_k(b^*))+\sum_{k=1}^{K}\alpha\|b_k-b_{k-1}\|^2. \nonumber
\end{align}
where $g_k(b) \triangleq f_k(b)+\beta L_k^*(b)$.
Recall that each service cost $f_k$ is $\theta_f$-strongly convex by Assumption 1. By Lemma~\ref{lem:Lk-convex}, $L_k^*(b)$ is convex in $b$. Since the sum of a strongly convex function and a convex function preserves strong convexity, the induced upper-level loss $g_k(b)$ is $\theta_g$-strongly convex over $\mathcal B=[B_0,T]$ with $\theta_g = \theta_f$.

Since the true transition model is unknown, the algorithm operates on a surrogate CMDP induced by the extended LP. Let $\hat L_k(b) \triangleq V_{\bar l} ({\pi_k};P_k)$ denote the optimistic surrogate value function. Similarly, $\hat L_k(b)$ is convex. Define the surrogate objective as $\hat g_k(b) \triangleq f_k(b)+ \beta \hat L_k(b)$.

The per-episode upper-level regret can be decomposed as follows:
(i) for the warm-up phase $1\le k\le K_0$, the regret is $g_k(B_0)-g_k(b^*)$;
(ii) for the online learning phase $k > K_0$, the regret term (including switching cost) is:
\begin{align*}
& g_k(b_k) - g_k(b^*)+\alpha\|b_k-b_{k-1}\|^2 \\
&= \underbrace{\hat g_k(b_k) - \hat g_k(b^*)+\alpha\|b_k-b_{k-1}\|^2}_{\text{Surrogate OCO Regret}} \\
&\quad + \underbrace{\beta \big( L(b_k) - \hat L(b_k) \big) - \beta \big( L(b^*) - \hat L(b^*) \big)}_{\text{Approximation Error}}.
\end{align*}

For $k > K_0$, the projected online subgradient update is
\[b_{k+1} = \Pi_{\gB}\bigl(b_k - \eta_k \hat h_k\bigr)\]
where $\hat h_k=\nabla f_k(b_k)-\beta\lambda_k$ and $\eta_k = \frac{1}{\theta_g (k-K_0)}$, here, $\lambda_k$ is the optimal dual multiplier associated with the budget constraint in the lower-level Extended LP (\ref{eq:ext-LP-constraint}) at episode $k$. To establish the boundedness of the subgradient $\hat h_k$, we rely on the convexity and subgradient of the value function $\hat L_k(b)$ (Lemmas \ref{lem:Lk-convex} and \ref{lem:Lk-subgradient}). Utilizing the safe baseline policy with $b_a < B_0$, and the convex function subgradient inequality $\hat L_k(b_a)-\hat L_k(b_k) \ge   - \lambda_k(b_a - b_k)$, we obtain the bound:$$\lambda_k \le \frac{\hat L_k(b_a) - \hat L_k(b_k)}{b_k - b_a} < \frac{T}{\gamma}.$$ 

The feasible region $\mathcal{B}$ has diameter $D=T-B_0$. Since $\|\nabla f_k(b)\|$ is bounded by $F$, and $\|\hat{h}_k\| = \|\nabla f_k - \beta \lambda_k\| \le \|\nabla f_k\| + \|\beta \lambda_k\|$ there exists a constant $G\triangleq  F + \beta \frac{T}{\gamma}$ such that the surrogate subgradient is bounded by $\|\hat{h}_k\| \le G$.

By the $\theta_g$-strong convexity of $\hat{g}_k$, we have:
\begin{align}
 \hat g_k(b_k) -\hat g_k(b^*)
  &\le \langle \hat h_k,\, b_k - b^*\rangle
     - \frac{\theta_g}{2}\,\|b_k - b^*\|^2. \label{eq:scvx}
\end{align}
Using the update rule for $b_{k+1}$ and the property of Euclidean projection:
\begin{align}
  \|b_{k+1}-b^*\|^2 &=\|\Pi_{[B_0,T]}\bigl(b_k - \eta_k \hat h_k\bigr)-b^*\|^2 \nonumber\\
  &\le \|b_k - \eta_k \hat h_k - b^*\|^2 \nonumber \\
  &= \|b_k - b^*\|^2 - 2\eta_k \langle \hat h_k, b_k - b^*\rangle + \eta_k^2 \|\hat h_k\|^2. \nonumber
\end{align}
Rearranging the terms yields:
\begin{align}
  \langle \hat h_k, b_k - b^*\rangle
  &\le
   \frac{1}{2\eta_k}\bigl(\|b_k-b^*\|^2 - \|b_{k+1}-b^*\|^2\bigr)
   + \frac{\eta_k}{2}\|\hat h_k\|^2. \label{eq:hat-inner}
\end{align}
Substituting (\ref{eq:hat-inner}) into (\ref{eq:scvx}) and using $\|\hat h_k\|\le G$, $\|b_k-b^*\|\le D$, we obtain:
\begin{align}
 \hat g_k(b_k) - \hat g_k(b^*)
  &\le
   \frac{1}{2\eta_k}\bigl(\|b_k-b^*\|^2 - \|b_{k+1}-b^*\|^2\bigr)\nonumber \\
  & \quad + \frac{\eta_k G^2}{2}
   - \frac{\theta_g}{2}\|b_k-b^*\|^2. \nonumber
\end{align}
For the switching cost:
\begin{align}
 \|b_{k}-b_{k-1}\|
&= \|\Pi_{[B_0,T]}(b_{k-1}-\eta_{k-1}\hat h_{k-1})- b_{k-1}\|\nonumber\\
&\le \|\eta_{k-1} \hat h_{k-1}\|
\le G\eta_{k-1}.\nonumber
\end{align}

Summing from $k=1$ to $K$, and setting $\eta_{K_0}=0$, we obtain:
\begin{align}
  & \mathrm{Reg}_{\mathrm{UL}}(K,b)
  = \sum_{k=1}^K \bigl(g_k(b_k) - g_k(b^*)\bigr)+\sum_{k=1}^{K}\alpha\|b_k-b_{k-1}\|^2\nonumber \\
  &=\sum_{k=1}^{K_0}(g_k(B_0)-g_k(b^*)) \nonumber \\
  &\quad+\sum_{k=K_0+1}^{K}(g_k(b_k)-g_k(b^*))+\sum_{k=K_0+1}^{K}\alpha\|b_k-b_{k-1}\|^2\nonumber\\
  &\le
   K_0 G D+\sum_{k=K_0+1}^K
   \left(\frac{1}{2\eta_k}-\frac{1}{2\eta_{k-1}}-\frac{\theta_g}{2}\right)\|b_k-b^*\|^2
   \nonumber\\
  &\quad
   + \sum_{k=K_0+1}^K \frac{\eta_k G^2}{2}+\sum_{k=K_0+1}^K (G\eta_k)^2 \nonumber\\
   &\quad + \sum_{k=K_0}^K \beta\big(( L_k^*(b_k) - \hat L_k(b_k))
 - ( L_k^*(b^*) - \hat L_k(b^*)) \big).  \label{eq:reg-sum-raw}
\end{align}

With the step size $\eta_k=\frac{1}{\theta_g(k-K_0)}$, the following bounds hold:
\begin{align} \label{1}
  \sum_{k=K_0+1}^K \frac{\eta_k G^2}{2}=\sum_{k=1}^{K-K_0-1} \frac{G^2}{2\theta_gk}
  &\le \frac{G^2}{2\theta_g}(1+\log K). 
\end{align}
\begin{align}
 \sum_{k=K_0+1}^K
   (\frac{1}{2\eta_k}-\frac{1}{2\eta_{k-1}}-\frac{\theta_g}{2})\bigl(\|b_k-b^*\|^2 \bigr)=0. 
\end{align}
\begin{align}
 \sum_{k=K_0+1}^{K}(G\eta_k)^2=G^2 \sum_{k=1}^{K-K_0} \frac{1}{\theta_g^2 k^2}
=
\frac{G^2 \pi^2}{6 \theta_g^2}.   
\end{align}
For the term $L(b_k)-\hat L(b_k)$, 
as $L(b_k)=V_{\bar l}({\pi_k^{\ac}};P_k)$, $\hat L(b_k)=V_{l}({\pi_k^{\ac^*}};P)$, then by lemma~\ref{lem:optimism}, we can get $V_{\bar l}({\pi_k^{\ac}};P_k)\le V_{l}({\pi_k^{\ac^*}};P)$. By (\ref{tilde}) and  (\ref{final_bound}), we can get
\[\sum_{k=K_0+1}^K V_l({\pi_k^{\ac}};P)-V_{\bar l}({\pi_k^{\ac}};P_k) \le \Tilde{O}\!\left(\frac{T^3}{\gamma}S \sqrt{AK} \right) \]
Consequently, we obtain:
\begin{align}
\sum_{k=K_0+1}^K\big(L(b_k)-\hat L(b_k)\big)&=\sum_{k=K_0+1}^K \big( V_l({\pi_k^{\ac^*}};P)-V_{\bar l}({\pi_k^{\ac}};P_k)\big) \nonumber\\
&\le \Tilde{O}\!\left(\frac{T^3}{\gamma}S \sqrt{AK} \right). \label{reggorate}   
\end{align}
Since $L(b_k)\ge L(b^*)$, it follows that:
\begin{align} \label{reggorate_optimal}
\sum_{k=K_0+1}^K\big(L(b^*)-\hat L(b^*)\big)\le \Tilde{O}\!\left(\frac{T^3}{\gamma}S \sqrt{AK} \right).
\end{align}

Adding  (\ref{1}) –  (\ref{reggorate_optimal}) to (\ref{eq:reg-sum-raw}), we obtain
a bound of the form:
\begin{align}
  \mathrm{Reg}_{\mathrm{UL}}(K,b)
  &\le
   \,\frac{G^2}{2\theta_g}\bigl(1+\log K\bigr)+ \Tilde{O}\!\left(\frac{T^3}{\gamma}S \sqrt{AK} \right) \nonumber\\
   &\quad+K_0GD+\frac{G^2 \pi^2}{6 \theta_g^2} 
  . \label{eq:reg-final}
\end{align}
\end{proof}
\section{Proof of Theorem~\ref{thm:total_regret}}\label{total_regret_1}

\begin{proof}
Recall that the total regret after $K$ episodes is defined as \[\regr(K, b,\pi) \triangleq \sum\nolimits_{k=1}^{K} C_k(b_k,\pi_k^{\ac}) - \sum\nolimits_{k=1}^{K} C_k(b^{*},\pi^{{\ac},*}).\] 
For each episode $k$ and a fixed upper-level decision $b_k$, define the \emph{episode-wise optimal safe lower-level policy} under the true model $P$:
\begin{align}
\pi_k^{*}(b_k) \in \argmin\nolimits_{\pi\in\Pi(b_k)} V_l(\pi;P),
\label{eq:lower_oracle}
\end{align}
where $\Pi(b_k)$ denotes the feasible policy set induced by budget $b_k$. Using $\pi_k^{*}(b_k)$ as an intermediate comparator, we can decompose the total regret as follows:
\begin{align}
&\regr(K, b,\pi) = \underbrace{\sum\nolimits_{k=1}^{K} \Big( C_k(b_k,\pi_k^{\ac}) - C_k(b_k,\pi_k^{*}(b_k)) \Big)}_{\text{(I) Lower-level BALDE regret under budget $b_k$: } \regr_{\mathrm{LL}}(K, \pi)} \nonumber \\
&\quad + \underbrace{\sum\nolimits_{k=1}^{K} \Big( C_k(b_k,\pi_k^{*}(b_k)) - C_k(b^{*},\pi^{{\ac},*}) \Big)}_{\text{(II) Upper-level BLOL regret with switching costs: } \regr_{\mathrm{UL}}(K, b)}. \label{eq:regret_decomp_1}
\end{align}
By the the definition of $C_k$ in (\ref{eq:episode_total_cost}), Term (I) in (\ref{eq:regret_decomp_1}) simplifies to:
\begin{align}
  &\sum\nolimits_{k=1}^{K} \Big( C_k(b_k,\pi_k^{\ac}) - C_k(b_k,\pi_k^{*}(b_k)) \Big)\nonumber\\
  &= \sum\nolimits_{k=1}^{K} \Big([
      f_k(b_k)
      + \alpha \bigl\|b_k - b_{k-1}\bigr\|^2
      + \beta V_l\bigl(\pi_k^{\mathrm{A}};P\bigr)
    ]
    \nonumber\\
  &-
    [
      f_k(b_k)
      + \alpha \bigl\|b_k - b_{k-1}\bigr\|^2
      + \beta V_l\bigl(\pi_k^*(b_k);P\bigr)
    ]\big) \nonumber\\
  &= \beta\sum\nolimits_{k=1}^{K}\big(
      V_l\bigl(\pi_k^{\mathrm{A}};P\bigr)
      - V_l\bigl(\pi_k^*(b_k);P\bigr)
    \big) \nonumber\\
 &=\beta R(K).
 \label{eq:term-I}
\end{align}

Term (II) in (\ref{eq:regret_decomp_1}) represents the upper-level BLOL regret with switching costs, i.e., the provisioning regret. Using the oracle effective loss $g_k(b)\triangleq f_k(b)+\beta L_k^{*}(b)$, where $L_k^{*}(b) = \min\nolimits_{\pi\in\Pi(b)} V_l(\pi;P)$ (see (\ref{eq:Lstar-def})), the upper-level regret term $\regr_{\mathrm{UL}}(K,b)$ becomes:
\begin{align}
\sum\nolimits_{k=1}^K \left( g_k(b_k) - g_k(b^*) \right) + \alpha \sum\nolimits_{k=1}^K \|b_k - b_{k-1}\|^2.
\label{eq:oracle_cost_identity_1}
\end{align}

Combining
(\ref{eq:regret_decomp_1})--(\ref{eq:oracle_cost_identity_1}),
Lemma~\ref{thm:bdope_main}, and Lemma~\ref{lem:oco_regret},
we obtain
\begin{align}
\regr(K, b,\pi)
&=
\regr_{\mathrm{LL}}(K, \pi)
+
\regr_{\mathrm{UL}}(K, b)
\nonumber\\
&=
\beta R(K)
+
\regr_{\mathrm{UL}}(K, b)
\nonumber\\
&=
\tilde{\mathcal O}
\left(
\frac{S T^{3}\sqrt{A K}}{\gamma}
\right)+\tilde{\mathcal O}(\log K).
\end{align}
\end{proof}

\section{Proof of Theorem~\ref{safety_guarantee}} \label{safety_guarantee_1}
In this section, we provide a detailed proof of Theorem~\ref{safety_guarantee}.  
The theorem establishes that, conditioned on the good event $\mathcal{G}$, 
the safety constraint $V_d(\pi_k;P) \le b_k$ holds for all episodes 
under the BALDE algorithm.

\begin{proof}
The proof is conditioned on the good event $\mathcal{G}$. We analyze the two phases of the algorithm separately.

\textit{Case 1: Warm-up phase ($k \le K_0$).}
During this phase, the algorithm executes $\pi_k = \pi_{\mathrm{base}}$ and the budget is fixed at $b_k = B_0$.
Since the baseline is safe with respect to $B_0$, we have:
\begin{align}
V_d(\pi_k; P) = V_d(\pi_{\mathrm{base}}; P) = b_{\mathrm{base}} < B_0 = b_k.
\end{align}
Thus, the constraint is trivally satisfied.

\textit{Case 2: Online learning phase ($k > K_0$).}
By Proposition~\ref{prop:dop-feasible}, the DOP problem is feasible for $k > K_0$. Let $(\pi_k, P_k)$ be the solution.
We apply the Value Difference Lemma (Lemma~\ref{lem:value_difference}) to the constriant cost function $d$:
\begin{align}
&V_d({\pi_k};P) - V_d({\pi_k};P_k)\nonumber\\
&= \mathbb{E}_{\pi_k, P_k}\!\left[
\sum_{t=1}^{T}
\big((P_t-P_{t,k})(\cdot \mid s_{k,t},a_{k,t})\big)^\top
V_{d,t+1}^{\pi_k}(\cdot;P)
\right]. \nonumber
\end{align}
Using Hölder's inequality and the fact that $d \in [0,1]$ implies $\|V_{d,t+1}^{\pi_k}\|_\infty \le T$, we bound the inner term:
\begin{align}
&\left| \big((P_t-P_{t,k})(\cdot \mid s_{k,t},a_{k,t})\big)^\top V_{d,t+1}^{\pi_k} \right|\nonumber\\
&\le \|P_t(\cdot \mid s_{k,t},a_{k,t}) - P_{t,k}(\cdot \mid s_{k,t},a_{k,t})\|_1 \cdot \|V_{d,t+1}^{\pi_k}\|_\infty \nonumber \\
&\le \sum_{s'}\beta_{k,t}^{p}(s_{k,t},a_{k,t},s') \cdot T. \nonumber
\end{align}
Taking the expectation, we obtain:
\begin{align}
V_d({\pi_k};P) - V_d({\pi_k};P_k)
&\le T \, \mathbb{E}_{\pi_k, P_k}\!\left[
\sum_{t=1}^{T} \sum_{s'}\beta_{k,t}^{p}(s_{k,t},a_{k,t},s')
\right] \nonumber \\
&= \epsilon_k^{\pi_k}(P_k). \label{eq:cost_diff_bound}
\end{align}
Rearranging (\ref{eq:cost_diff_bound}), we have $V_d({\pi_k};P) \le V_d({\pi_k};P_k) + \epsilon_k^{\pi_k}(P_k)$.
Recall that the DOP problem imposes the pessimistic constraint $V_{\bar{d}}(\pi_k; P_k) \le b_k$. Using the decomposition property $V_{\bar{d}}(\pi; P') = V_d(\pi; P') + \epsilon_k^{\pi}(P')$, we have:
\begin{align}
V_d({\pi_k};P_k) + \epsilon_k^{\pi_k}(P_k) = V_{\bar{d}}(\pi_k; P_k) \le b_k.
\end{align}
Combining these inequalities yields:
\begin{align}
V_d({\pi_k};P) \le V_d({\pi_k};P_k) + \epsilon_k^{\pi_k}(P_k) \le b_k.
\end{align}
This confirms that the safety constraint is satisfied in the true environment for all $k > K_0$.

Combining Case 1 and Case 2, the constraint is never violated with probability at least $1-3\delta$.
\end{proof}